\documentclass[runningheads]{llncs}

\usepackage{eccv}

\usepackage{eccvabbrv}

\usepackage{graphicx}
\usepackage{booktabs}
\usepackage{bibunits}

\usepackage[accsupp]{axessibility}  

\usepackage[pagebackref,breaklinks,colorlinks,citecolor=eccvblue]{hyperref}

\usepackage{bm}
\usepackage{url}
\usepackage{multirow}
\usepackage{makecell}
\usepackage{graphicx}
\usepackage{amsmath}
\usepackage{amssymb}
\usepackage{booktabs}
\usepackage{algorithm}
\usepackage{algpseudocode}
\usepackage{pifont}
\usepackage{xcolor}
\definecolor{bluekeywords}{rgb}{0.13,0.13,1}
\newcommand{\FunctionName}[1]{\textcolor{bluekeywords}{\textsc{#1}}}
\newcommand{\blackdingB}[2]{\textcolor{#2}{\ding{\numexpr201+#1\relax}}}

\renewcommand\toprule{\specialrule{1.5pt}{1pt}{0pt}}
\renewcommand\bottomrule{\specialrule{1.5pt}{0pt}{1pt}}

\usepackage{marvosym} 

\usepackage[english]{babel}

\usepackage[pagebackref,breaklinks,colorlinks]{hyperref}

\usepackage{orcidlink}
\usepackage[capitalize]{cleveref}
\crefname{section}{Sec.}{Secs.}
\Crefname{section}{Section}{Sections}
\Crefname{table}{Table}{Tables}
\crefname{table}{Tab.}{Tabs.}

\begin{document}

\title{FouriScale: A Frequency Perspective on Training-Free High-Resolution Image Synthesis} 

\titlerunning{FouriScale}

\author{Linjiang Huang \inst{1,2}\thanks{Equal contribution. \quad \textsuperscript{\Letter} Corresponding author.} \and
Rongyao Fang \inst{1\star} \and
Aiping Zhang \inst{3} \and
Guanglu Song \inst{4} \\
Si Liu \inst{5} \and
Yu Liu \inst{4} \and
Hongsheng Li \inst{1,2} \textsuperscript{\Letter}}

\authorrunning{L.~Huang et al.}

\institute{CUHK-SenseTime Joint Laboratory, The Chinese University of Hong Kong \and
Centre for Perceptual and Interactive Intelligence \and
Sun Yat-Sen University \and
Sensetime Research \and Beihang University \\
\email{ljhuang524@gmail.com}, \quad \email{\{rongyaofang@link, hsli@ee\}.cuhk.edu.hk}
}

\maketitle

\begin{abstract}
In this study, we delve into the generation of high-resolution images from pre-trained diffusion models, addressing persistent challenges, such as repetitive patterns and structural distortions, that emerge when models are applied beyond their trained resolutions. To address this issue, we introduce an innovative, training-free approach FouriScale from the perspective of frequency domain analysis.
We replace the original convolutional layers in pre-trained diffusion models by incorporating a dilation technique along with a low-pass operation, intending to achieve structural consistency and scale consistency across resolutions, respectively. Further enhanced by a padding-then-crop strategy, our method can flexibly handle text-to-image generation of various aspect ratios. By using the FouriScale as guidance, our method successfully balances the structural integrity and fidelity of generated images, achieving an astonishing capacity of arbitrary-size, high-resolution, and high-quality generation. With its simplicity and compatibility, our method can provide valuable insights for future explorations into the synthesis of ultra-high-resolution images. The code will be released at \url{https://github.com/LeonHLJ/FouriScale}.
\end{abstract}

\section{Introduction} \label{sec:intro}

Recently, Diffusion models~\cite{ho2020denoising,rombach2022high} have emerged as the predominant generative models, surpassing the popularity of GANs~\cite{goodfellow2014generative} and autoregressive models~\cite{ramesh2021zero,ding2021cogview}. 
Some text-to-image generation models, which are based on diffusion models, such as Stable Diffusion (SD)~\cite{rombach2022high}, Stable Diffusion XL (SDXL)~\cite{podell2023sdxl}, Midjourney~\cite{midjourney}, and Imagen~\cite{saharia2022photorealistic}, have shown their astonishing capacity to generate high-quality and fidelity images under the guidance of text prompts. To ensure efficient processing on existing hardware and stable model training, these models are typically trained at one or a few specific image resolutions. For instance, SD models are often trained using images of $512 \times 512$ resolution, while SDXL models are typically trained with images close to $1024 \times 1024$ pixels.

\begin{figure}[!t]
  \centering
  \includegraphics[width=0.98\columnwidth]{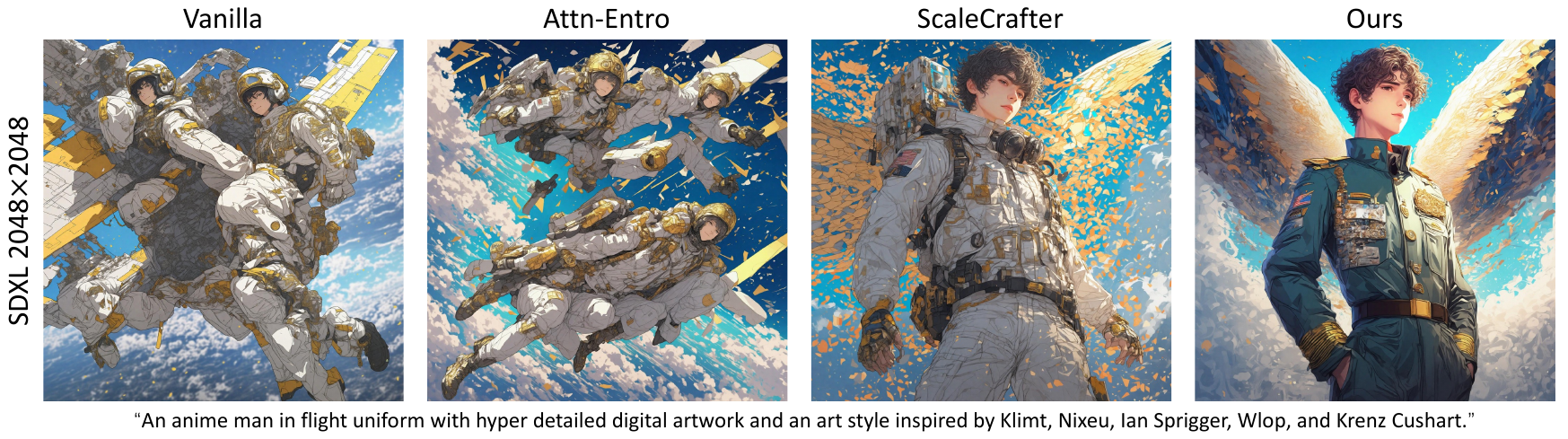}
  \caption{Visualization of pattern repetition issue of higher-resolution image synthesis using pre-trained SDXL~\cite{podell2023sdxl} (Train: 1024$\times$1024; Inference:2048$\times$2048). Attn-Entro~\cite{jin2023training} fails to address this problem and ScaleCrafter~\cite{he2023scalecrafter} still struggles with this issue in image details. Our method successfully handles this problem and generates high-quality images without model retraining.} \label{fig:visualization_1}
  \vspace{-3mm}
\end{figure}

However, as shown in Fig.~\ref{fig:visualization_1}, directly employing pre-trained diffusion models to generate an image at a resolution higher than what the models were trained on will lead to significant issues, including repetitive patterns and unforeseen artifacts. Some studies~\cite{bar2023multidiffusion, jimenez2023mixture,lee2024syncdiffusion} have attempted to create larger images by utilizing pre-trained diffusion models to stitch together overlapping patches into a panoramic image. Nonetheless, the absence of a global direction for the whole image restricts their ability to generate images focused on specific objects and fails to address the problem of repetitive patterns, where a unified global structure is essential. Recent work~\cite{jin2023training} has explored adapting pre-trained diffusion models for generating images of various sizes by examining attention entropy. Nevertheless, ScaleCrafter~\cite{he2023scalecrafter} found that the key point of generating high-resolution images lies in the convolution layers. They introduce a re-dilation operation and a convolution disperse operation to enlarge kernel sizes of convolution layers, largely mitigating the problem of pattern repetition. However, their conclusion stems from empirical findings, lacking a deeper exploration of this issue. Additionally, it needs an initial offline computation of a linear transformation between the original convolutional kernel and the enlarged kernel, falling short in terms of compatibility and scalability when there are variations in the kernel sizes of the UNet and the desired target resolution of images.

In this work, we present FouriScale, an innovative and effective approach that handles the issue through the perspective of frequency domain analysis, successfully demonstrating its effectiveness through both theoretical analysis and experimental results.
FouriScale substitutes the original convolutional layers in pre-trained diffusion models by simply introducing a dilation operation coupled with a low-pass operation, aimed at achieving structural and scale consistency across resolutions, respectively. Equipped with a padding-then-crop strategy, our method allows for flexible text-to-image generation of different sizes and aspect ratios. Furthermore, by utilizing FouriScale as guidance, our approach attains remarkable capability in producing high-resolution images of any size, with integrated image structure alongside superior quality. The simplicity of FouriScale eliminates the need for any offline pre-computation, facilitating compatibility and scalability. We envision FouriScale providing significant contributions to the advancement of ultra-high-resolution image synthesis in future research.

\section{Related Work} \label{sec:related_work}

\subsection{Text-to-Image Synthesis}

Text-to-image synthesis~\cite{dhariwal2021diffusion,ho2022cascaded,rombach2022high,saharia2022photorealistic} has seen a significant surge in interest due to the development of diffusion probabilistic models~\cite{ho2020denoising,song2020denoising}. These innovative models operate by generating data from a Gaussian distribution and refining it through a denoising process. With their capacity for high-quality generation, they have made significant leaps over traditional models like GANs~\cite{goodfellow2014generative,dhariwal2021diffusion}, especially in producing more realistic images.
The Latent Diffusion Model (LDM)~\cite{rombach2022high} integrates the diffusion process within a latent space, achieving astonishing realism in the generation of images, which boosts significant interest in the domain of generating via latent space~\cite{he2022latent,zeng2022lion,peebles2023scalable,blattmann2023align,liu2023audioldm}. To ensure efficient processing on existing hardware and stable model training, these models are typically trained at one or a few specific image resolutions. For instance, Stabe Diffusion (SD)~\cite{rombach2022high} is trained using $512 \times 512$ pixel images, while SDXL~\cite{podell2023sdxl} models are typically trained with images close to $1024 \times 1024$ resolution, accommodating various aspect ratios simultaneously.

\subsection{High-Resolution Synthesis via Diffusion Models}

High-resolution synthesis has always received widespread attention. Prior works mainly focus on refining the noise schedule~\cite{chen2023importance,hoogeboom2023simple}, developing cascaded architectures~\cite{ho2022cascaded,saharia2022photorealistic,teng2023relay} or mixtures-of-denoising-experts~\cite{balaji2022ediffi} for generating high-resolution images.
Despite their impressive capabilities, diffusion models were often limited by specific resolution constraints and did not generalize well across different aspect ratios and resolutions. Some methods have tried to address these issues by accommodating a broader range of resolutions. For example, Any-size Diffusion~\cite{zheng2023any} fine-tunes a pre-trained SD on a set of images with a fixed range of aspect ratios, similar to SDXL~\cite{podell2023sdxl}.
FiT~\cite{lu2024fit} views the image as a sequence of tokens and adaptively padding image tokens to a predefined maximum token limit, ensuring hardware-friendly training and flexible resolution handling. However, these models require model training, overlooking the inherent capability of the pre-trained models to handle image generation with varying resolutions. Most recently, some methods~\cite{bar2023multidiffusion, jimenez2023mixture,lee2024syncdiffusion} have attempted to generate panoramic images by utilizing pre-trained diffusion models to stitch together overlapping patches. Recent work~\cite{jin2023training} has explored adapting pre-trained diffusion models for generating images of various sizes by examining attention entropy. ElasticDiff~\cite{haji2023elasticdiffusion} uses the estimation of default resolution to guide the generation of arbitrary-size images. However, ScaleCrafter~\cite{he2023scalecrafter} finds that the key point of generating high-resolution images by pre-trained diffusion models lies in convolution layers. They present a re-dilation and a convolution disperse operation to expand convolution kernel sizes, which requires an offline calculation of a linear transformation from the original convolutional kernel to the expanded one. In contrast, we deeply investigate the issue of repetitive patterns and handle it through the perspective of frequency domain analysis. The simplicity of our method eliminates the need for any offline pre-computation, facilitating its compatibility and scalability.

\section{Method} \label{sec:proposed_method}

\begin{figure}[!t]
  \centering
  \includegraphics[width=0.98\columnwidth]{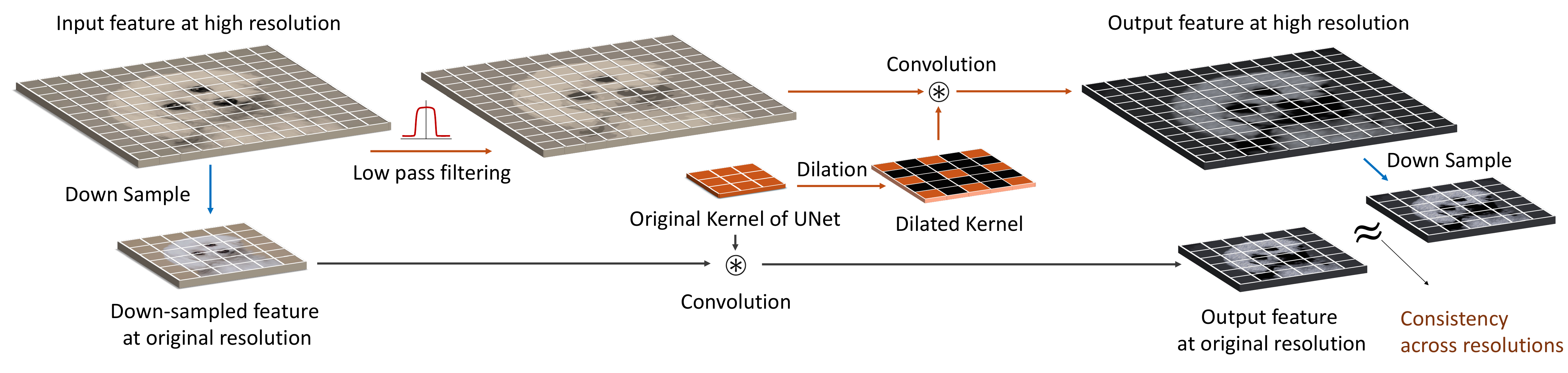}
  \caption{The overview of FouriScale (orange line), which includes a dilation convolution operation (Sec.~\ref{sec:dilated_conv}) and a low-pass filtering operation (Sec.~\ref{sec:low_pass}) to achieve structural consistency and scale consistency across resolutions, respectively.} \label{fig:fouriscale_overview}
\end{figure}

Diffusion models, also known as score-based generative models~\cite{ho2020denoising, song2020denoising}, belong to a category of generative models that follow a process of progressively introducing Gaussian noise into the data and subsequently generating samples from this noise through a reverse denoising procedure. The key denoising step is typically carried out by a U-shaped Network (UNet), which learns the underlying denoising function that maps from noisy data to its clean counterpart. The UNet architecture, widely adopted for this purpose, comprises stacked convolution layers, self-attention layers, and cross-attention layers. Some previous works have explored the degradation of performance when the generated resolution becomes larger, attributing to the change of the attention tokens' number~\cite{jin2023training} and the reduced relative receptive field of convolution layers~\cite{he2023scalecrafter}. Based on empirical evidence in~\cite{he2023scalecrafter}, convolutional layers are more sensitive to changes in resolution. Therefore, we primarily focus on studying the impact brought about by the convolutional layers. In this section, we will introduce FouriScale, as shown in Fig.~\ref{fig:fouriscale_overview}. It includes a dilation convolution operation (Sec.~\ref{sec:dilated_conv}) and a low-pass filtering operation (Sec.~\ref{sec:low_pass}) to achieve structural consistency and scale consistency across resolutions, respectively. With the tailored padding-then-cropping strategy (Sec.~\ref{sec:arbitrary_generation}), FouriScale can generate images of arbitrary aspect ratios. By utilizing FouriScale as guidance (Sec.~\ref{sec:fouri_guidance}), our approach attains remarkable capability in generating high-resolution and high-quality images.

\subsection{Notation} \label{sec:Preliminary}

\paragraph{2D Discrete Fourier Transform (2D DFT).}
Given a two-dimensional discrete signal $F(m, n)$ with dimensions $M \times N $, the two-dimensional discrete Fourier transform (2D DFT) is defined as:
\begin{equation} \label{eq:2D_DFT}
F(p, q) = \frac{1}{MN} \sum_{m=0}^{M-1} \sum_{n=0}^{N-1} F(m, n) e^{-j2\pi\left(\frac{pm}{M} + \frac{qn}{N}\right)}.
\end{equation}

\paragraph{2D Dilated Convolution.}
A dilated convolution kernel of the kernel $k(m, n)$, denoted as $k_{d_h, d_w}(m, n)$, is formed by introducing zeros between the elements of the original kernel such that:
\begin{equation}
k_{d_h, d_w}(m, n) =
\begin{cases} 
k(\frac{m}{d_h}, \frac{n}{d_w}) & \text{if } m \operatorname{\%} d_h = 0 \text{ and } n \operatorname{\%} d_w = 0, \\
0 & \text{otherwise},
\end{cases} 
\end{equation}
where $d_h$, $d_w$ is the dilation factor along height and width, respectively, $m$ and $n$ are the indices in the dilated space. The $\%$  represents the modulo operation.

\setcounter{footnote}{0}
\subsection{Structural Consistency via Dilated Convolution} \label{sec:dilated_conv}

The diffusion model's denoising network, denoted as $\epsilon_{\theta}$, is generally trained on images or latent spaces at a specific resolution of $h\times w$. This network is often constructed using a U-Net architecture. Our target is to generate an image of a larger resolution of $H\times W$ at the inference stage using the parameters of denoising network $\epsilon_{\theta}$ without retraining.

As previously discussed, the convolutional layers within the U-Net are largely responsible for the occurrence of pattern repetition when the inference resolution becomes larger. To prevent structural distortion at the inference resolution, we resort to establishing structural consistency between the default resolution and high resolutions, as shown in Fig.~\ref{fig:fouriscale_overview}. In particular, for a convolutional layer $\operatorname{Conv}_k$ in the UNet with its convolution kernel $k$, and the high-resolution input feature map $F$, the structural consistency can be formulated as follows:
\begin{equation} \label{eq:structural_consistency}
\operatorname{Down}_{s}(F) \circledast k = \operatorname{Down}_{s}(F \circledast k'),
\end{equation}
where $\operatorname{Down}_{s}$ denotes the down-sampling operation with scale $s$\footnote{For simplicity, we assume equal down-sampling scales for height and width. Our method can also accommodate different down-sampling scales in this context through our padding-then-cropping strategy (Section \ref{sec:arbitrary_generation}).}, and $\circledast$ represents the convolution operation. This equation implies the need to customize a new convolution kernel $k'$ for a larger resolution. However, finding an appropriate $k'$ can be challenging due to the variety of feature map $F$. The recent ScaleCrafter~\cite{he2023scalecrafter} method uses structure-level and pixel-level calibrations to learn a linear transformation between $k$ and $k'$, but learning a new transformation for each new kernel size and new target resolution can be cumbersome.

In this work, we propose to handle the structural consistency from a frequency perspective. Suppose the input $F(x,y)$, which is a two-dimensional discrete spatial signal, belongs to the set $\mathbb{R}^{H_f \times W_f \times C}$. The sampling rates along the $x$ and $y$ axes are given by $\Omega_x$ and $\Omega_y$ correspondingly. The Fourier transform of $F(x, y)$ is represented by $F(u,v) \in \mathbb{R}^{H_f \times W_f \times C}$. In this context, the highest frequencies along the $u$ and $v$ axes are denoted as $u_{max}$ and $v_{max}$, respectively. Additionally, the Fourier transform of the downsampled feature map $\operatorname{Down}_{s}(F(x, y))$, which is dimensionally reduced to $\mathbb{R}^{ \frac{H_f}{s} \times \frac{W_f}{s} \times C}$, is denoted as $F'(u,v)$. 

\begin{theorem} \label{theo:1d_signal}
Spatial down-sampling leads to a reduction in the range of frequencies that the signal can accommodate, particularly at the higher end of the spectrum. This process causes high frequencies to be folded to low frequencies, and superpose onto the original low frequencies. For a one-dimensional signal, in the condition of $s$ strides, this superposition of high and low frequencies resulting from down-sampling can be mathematically formulated as
\begin{equation}
F'(u) = \mathbb{S}(F(u), F\left(u + \frac{a \Omega_x}{s}\right)) \mid  u \in \left(0, \frac{\Omega_x}{s}\right),
\end{equation}
where $\mathbb{S}$ dentes the superposing operator, $\Omega_x$ is the sampling rates in $x$ axis, and $a = 1, \ldots, s-1$.
\end{theorem}

\begin{lemma} \label{lemma:image_superpose}
For an image, the operation of spatial down-sampling using strides of $s$ can be viewed as partitioning the Fourier spectrum into $s \times s$ equal patches and then uniformly superimposing these patches with an average scaling of $\frac{1}{s^2}$.
\begin{equation} \label{eq:superpose}
\operatorname{DFT}\left(\operatorname{Down}_{s} (F(x, y))\right) = \frac{1}{s^2} \sum_{i=0}^{s-1} \sum_{j=0}^{s-1} F_{(i,j)}(u, v),
\end{equation}
where $F_{(i,j)}(u, v)$ is a sub-matrix of $F(u, v)$ by equally splitting $F(u, v)$ into $s \times s$ non-overlapped patches and $i, j \in \{0, 1, \ldots, s-1 \}$.
\end{lemma}
\begin{figure}[!t]
  \centering
  \includegraphics[width=0.98\columnwidth]{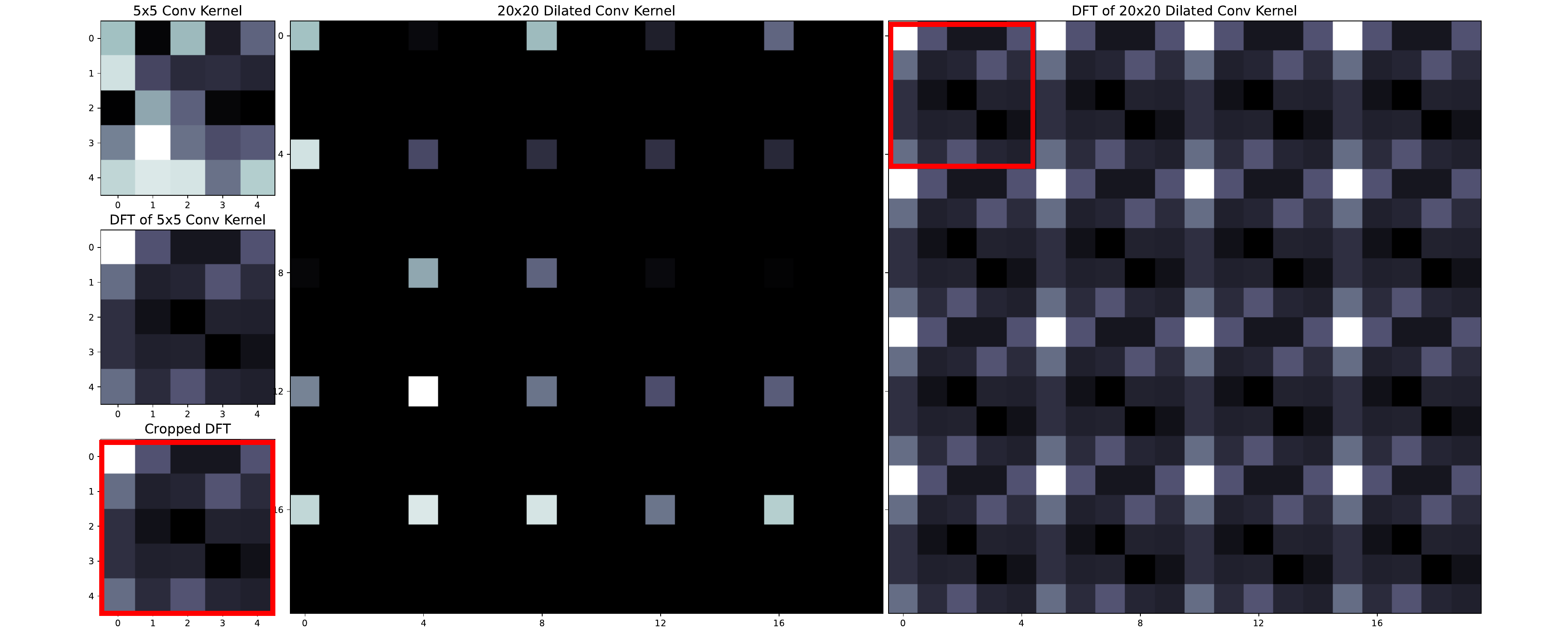}
  \caption{We visualize a random $5 \times 5$ kernel for better visualization. The Fourier spectrum of its dilated kernel, with a dilation factor of 4, clearly demonstrates a periodic character. It should be noted that we also pad zeros to the right and bottom sides of the dilated kernel, which differs from the conventional use. However, this does not impact the outcome in practical applications.} \label{fig:dilation_freq_preodic_vis}
  \vspace{-2mm}
\end{figure}

The proof of Theorem~\ref{theo:1d_signal} and Lemma~\ref{lemma:image_superpose} are provided in the Appendix (Sec.~\ref{sec:proof_theorem} and Sec.~\ref{sec:proof_lemma}). They describe the shuffling and superposing~\cite{zhang2019making,riad2021learning,zhu2023fouridown} in the frequency domain imposed by spatial down-sampling. If we transform Eq.~\eqref{eq:structural_consistency} to the frequency domain and follow conclusion in Lemma~\ref{lemma:image_superpose}, we can obtain:
\begin{flalign} \label{eq:eq_to_frequency_domain}
&\left( \frac{1}{s^2} \sum_{i=0}^{s-1} \sum_{j=0}^{s-1} F_{(i,j)}(u, v) \right) \odot k(u, v) \leftarrow \text{Left side of Eq.~\eqref{eq:structural_consistency}} \nonumber \\
&= \frac{1}{s^2} \sum_{i=0}^{s-1} \sum_{j=0}^{s-1} \left(F_{(i,j)}(u, v) \odot k(u, v) \right) \\
&= \frac{1}{s^2} \sum_{i=0}^{s-1} \sum_{j=0}^{s-1} \left(F_{(i,j)}(u, v) \odot k'_{(i,j)}(u, v) \right), \leftarrow \text{Right side of Eq.~\eqref{eq:structural_consistency}} \nonumber
\end{flalign}
where $k(u, v)$, $k'(u, v)$ denote the fourier transform of kernel $k$ and $k'$, respectively, $\odot$ is element-wise multiplication. Eq.~\eqref{eq:eq_to_frequency_domain} suggests that the Fourier spectrum of the ideal convolution kernel $k'$ should be the one that is stitched by $s \times s$ Fourier spectrum of the convolution kernel $k$. In other words, there should be a periodic repetition in the Fourier spectrum of $k'$, the repetitive pattern is the Fourier spectrum of $k$.

Fortunately, the widely used dilated convolution perfectly meets this requirement. 
Suppose a kernel $k(m,n)$ with the size of $M \times N$, it's dilated version is $k_{d_h, d_w}(m,n)$, with dilation factor of $(d_h, d_w)$. For any integer multiples of $d_h$, namely \( p' = pd_h \) and integer multiples of $d_w$, namely \( q' = qd_w \), the exponential term of the dilated kernel in the 2D DFT (Eq.~\eqref{eq:2D_DFT}) becomes:
\begin{equation}
e^{-j2\pi\left(\frac{p'm}{d_hM} + \frac{q'n}{d_wN}\right)} = e^{-j2\pi\left(\frac{pm}{M} + \frac{qn}{N}\right)},
\end{equation}
which is periodic with a period of $M$ along the $m$-dimension and a period of $N$ along the $n$-dimension. It indicates that a dilated convolution kernel parameterized by the original kernel $k$, with dilation factor of $(H/h, W/w)$, is the ideal convolution kernel $k'$. In Fig.~\ref{fig:dilation_freq_preodic_vis}, we visually demonstrate the periodic repetition of dilated convolution. We noticed that~\cite{he2023scalecrafter} also uses dilated operation. In contrast to~\cite{he2023scalecrafter}, which is from empirical observation, our work begins with a focus on frequency analysis and provides theoretical justification for its effectiveness.

\subsection{Scale Consistency via Low-pass Filtering} \label{sec:low_pass}
\begin{figure}[!t]
    \centering
    \begin{subfigure}[b]{0.27\linewidth}
        \includegraphics[width=\linewidth]{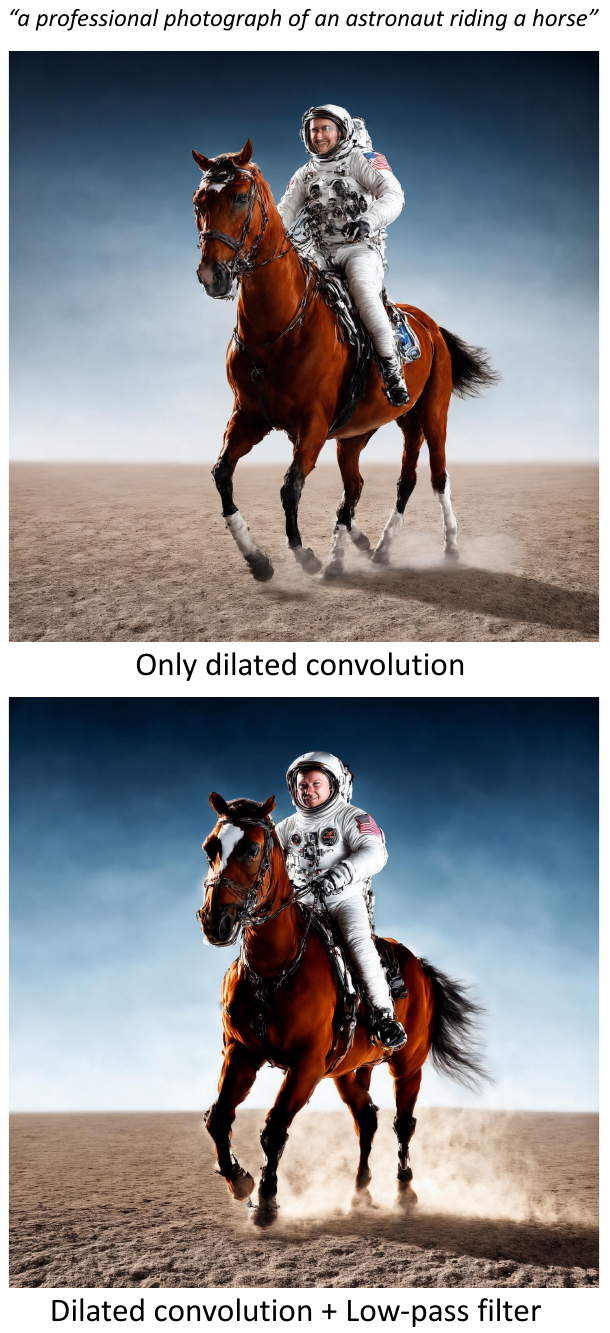}
        \caption{Visual comparisons}
        \label{fig:only_dilated}
    \end{subfigure}
    \hfill
    \begin{subfigure}[b]{0.35\linewidth}
        \includegraphics[width=\linewidth]{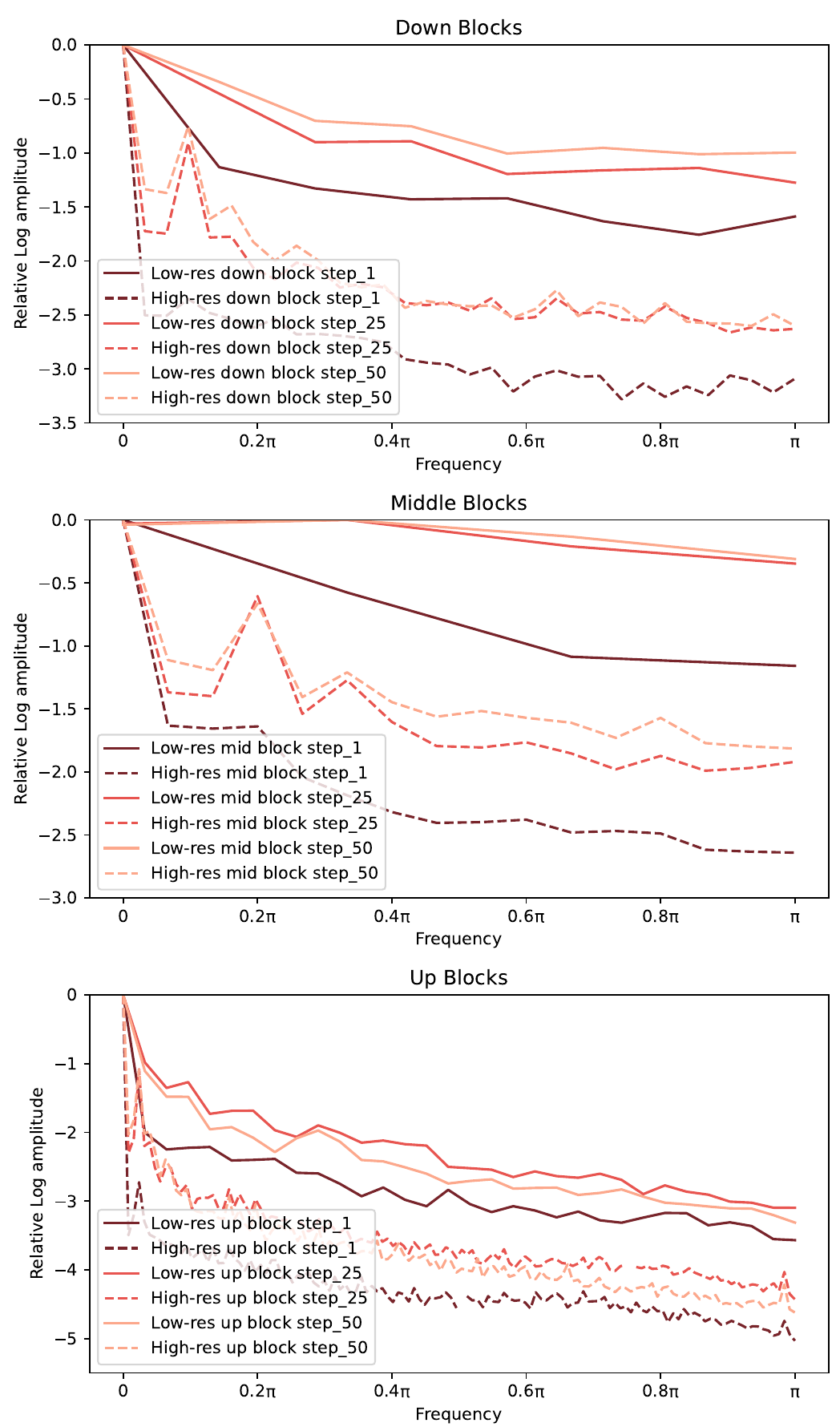}
        \caption{without filtering}
        \label{fig:without_filtering}
    \end{subfigure}
    \hfill
    \begin{subfigure}[b]{0.35\linewidth}
        \includegraphics[width=\linewidth]{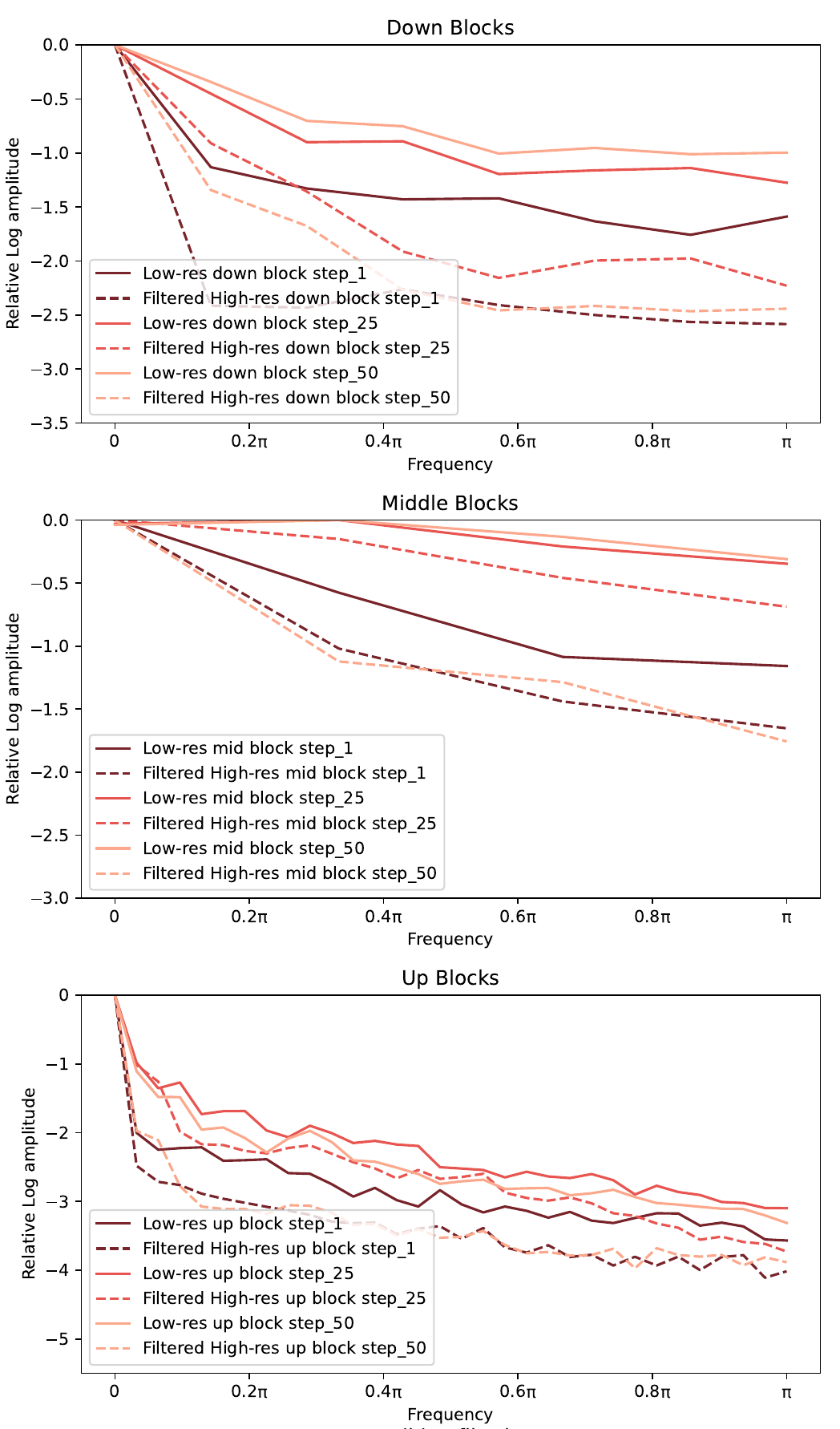}
        \caption{with filtering}
        \label{fig:with_filtering}
    \end{subfigure}
    \caption{(a) Visual comparisons between the images created at a resolution of $2048 \times 2048$: with only the dilated convolution, and with both the dilated convolution and the low-pass filtering. (b)(c) Fourier relative log amplitudes of input features from three distinct layers from the down blocks, mid blocks, and up blocks of UNet, respectively, are analyzed. We also include features at reverse steps 1, 25, and 50. (b) Without the application of the low-pass filter. There is an evident distribution gap of the frequency spectrum between the low resolution and high resolution. (c) With the application of the low-pass filter. The distribution gap is largely reduced.} \label{fig:vis_Fourier_relative_log_amplitudes}
    \vspace{-1mm}
\end{figure}

However, in practice, dilated convolution alone cannot well mitigate the issue of pattern repetition. As shown in Fig.~\ref{fig:only_dilated} (top left), the issue of pattern repetition is significantly reduced, but certain fine details, like the horse's legs, still present issues. This phenomenon is because of the aliasing effect after the spatial down-sampling, which raises the distribution gap between the features of low resolution and the features down-sampled from high resolution, as presented in Fig.~\ref{fig:without_filtering}. Aliasing alters the fundamental frequency components of the original signal, breaking its consistency across scales.

In this paper, we introduce a low-pass filtering operation, or spectral pooling~\cite{rippel2015spectral} to remove high-frequency components that might cause aliasing, intending to construct scale consistency across different resolutions. Let $F(m, n)$ be a two-dimensional discrete signal with resolution \(M \times N\). Spatial down-sampling of $F(m, n)$, by factors \(s_h\) and \(s_w\) along the height and width respectively, alters the Nyquist limits to \(M/(2s_h)\) and \(N/(2s_w)\) in the frequency domain, corresponding to half the new sampling rates along each dimension.
The expected low-pass filter should remove frequencies above these new Nyquist limits to prevent aliasing. 
Therefore, the optimal mask size (assuming the frequency spectrum is centralized) for passing low frequencies in a low-pass filter is \(M/s_h \times N/s_w\).
This filter design ensures the preservation of all valuable frequencies within the downscaled resolution while preventing aliasing by filtering out higher frequencies.

As illustrated in Fig.~\ref{fig:with_filtering}, the application of the low-pass filter results in a closer alignment of the frequency distribution between high and low resolutions. This ensures that the left side of Eq.~\eqref{eq:structural_consistency} produces a plausible image structure.
Additionally, since our target is to rectify the image structure, low-pass filtering would not be harmful because it generally preserves the structural information of a signal, which predominantly resides in the lower frequency components~\cite{pattichis2007analyzing,zhang2018image}.

Subsequently, the final kernel $k^*$ is obtained by applying low-pass filtering to the dilated kernel. Considering the periodic nature of the Fourier spectrum associated with the dilated kernel, the Fourier spectrum of the new kernel $k^*$ involves expanding the spectrum of the original kernel $k$ by inserting zero frequencies. Therefore, this expansion avoids the introduction of new frequency components into the new kernel $k^*$. In practice, we do not directly calculate the kernel $k^*$ but replace the original $\operatorname{Conv}_k$ with the following equivalent operation to ensure computational efficiency:
\begin{equation} \label{eq:final_operation}
\operatorname{Conv}_k(F) \rightarrow \operatorname{Conv}_{k'}(\operatorname{iDFT}(H \odot \operatorname{DFT}(F)),
\end{equation}
where $H$ denotes the low-pass filter. Fig.~\ref{fig:only_dilated} (bottom left) illustrates that the combination of dilated convolution and low-pass filtering resolves the issue of pattern repetition.

\subsection{Adaption to Arbitrary-size Generation} \label{sec:arbitrary_generation}
\begin{figure}[!t]
  \centering
  \includegraphics[width=0.98\columnwidth]{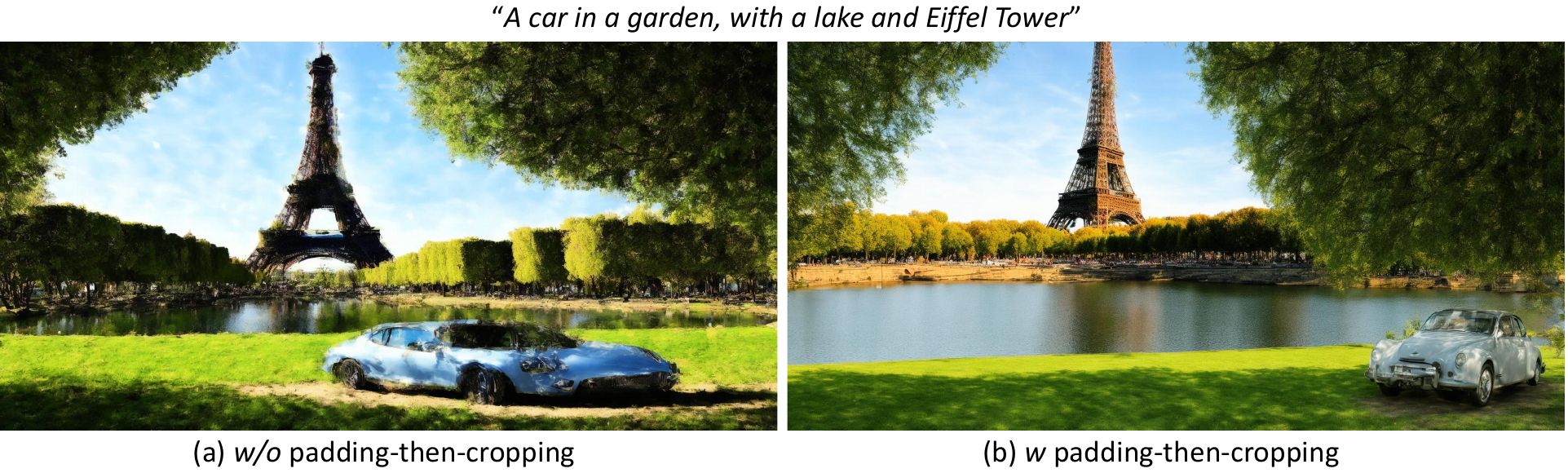}
  \caption{Visual comparisons between the images created at a resolution of $2048 \times 1024$: (a) without the application of padding-then-cropping strategy, and (b) with the application of padding-then-cropping strategy. The Stable Diffusion 2.1 utilized is initially trained on images of $512 \times 512$ resolution.} \label{fig:padding-then-cropping}
\end{figure}

The derived conclusion is applicable only when the aspect ratios of the high-resolution image and the low-resolution image used in training are identical. 
From Eq.~\eqref{eq:superpose} and Eq.~\eqref{eq:eq_to_frequency_domain}, it becomes apparent that when the aspect ratios vary, meaning the dilation rates along the height and width are different, the well-constructed structure in the low-resolution image would be distorted and compressed, as shown in Fig.~\ref{fig:padding-then-cropping} (a).
Nonetheless, in real-world applications, the ideal scenario is for a pre-trained diffusion model to have the capability of generating arbitrary-size images.
\begin{algorithm}[!t]
\caption{Pseudo-code of FouriScale} \label{algo:final_operation}
\begin{algorithmic}
\State \textbf{Data:} Input: $F \in \mathbb{R}^{C \times H_f \times W_f}$. Original size: $h_f \times w_f$.
\State \textbf{Result:} Output: $F_{conv} \in \mathbb{R}^{C \times H_f \times W_f}$
\State $r = max(\lceil \frac{H_f}{h_f} \rceil, \lceil \frac{W_f}{w_f} \rceil)$ 
\State $F_{pad} \gets \FunctionName{Zero-Pad}(F) \in \mathbb{R}^{C \times rh_f \times rw_f}$ \Comment{Zero Padding}
\State $F_{dft} \gets \FunctionName{DFT}(F_{pad}) \in \mathbb{C}^{C \times rh_f \times rw_f}$ \Comment{Discrete Fourier transform}
\State $F_{low} \gets H \odot F_{dft}$ \Comment{Low pass filtering}
\State $F_{idft} \gets \FunctionName{iDFT}(F_{low})$ \Comment{Inverse Fourier transform}
\State $F_{crop} \gets \FunctionName{Crop}(F_{idft}) \in \mathbb{R}^{R \times H_f \times W_f}$ \Comment{Cropping}
\State $F_{conv} \gets \FunctionName{Conv}_{k'}(F_{crop})$ \Comment{Dilation factor of $k'$ is $r$}
\end{algorithmic}
\end{algorithm}

We introduce a straightforward yet efficient approach, termed \emph{padding-then-cropping}, to solve this problem. Fig.~\ref{fig:padding-then-cropping} (b) demonstrates its effectiveness. In essence, when a layer receives an input feature at a standard resolution of $h_f \times w_f$, and this input feature increases to a size of $H_f \times W_f $ during inference, our first step is to zero-pad the input feature to a size of $ r h_f \times r w_f$. Here, $r$ is defined as the maximum of $\lceil \frac{H_f}{h_f} \rceil$ and $\lceil \frac{W_f}{w_f} \rceil$, with $\lceil \cdot \rceil$ representing the ceiling operation. The padding operation assumes that we aim to generate an image of size $ r h \times r w$, where certain areas are filled with zeros.
Subsequently, we apply Eq.~\eqref{eq:final_operation} to rectify the issue of repetitive patterns in the higher-resolution output.
Ultimately, the obtained feature is cropped to restore its intended spatial size. This step is necessary to not only negate the effects of zero-padding but also control the computational demands when the resolution increases, particularly those arising from the self-attention layers in the UNet architecture. 
Taking computational efficiency into account, our equivalent solution is outlined in Algorithm \ref{algo:final_operation}.
\begin{figure}[!t]
  \centering
  \includegraphics[width=0.98\columnwidth]{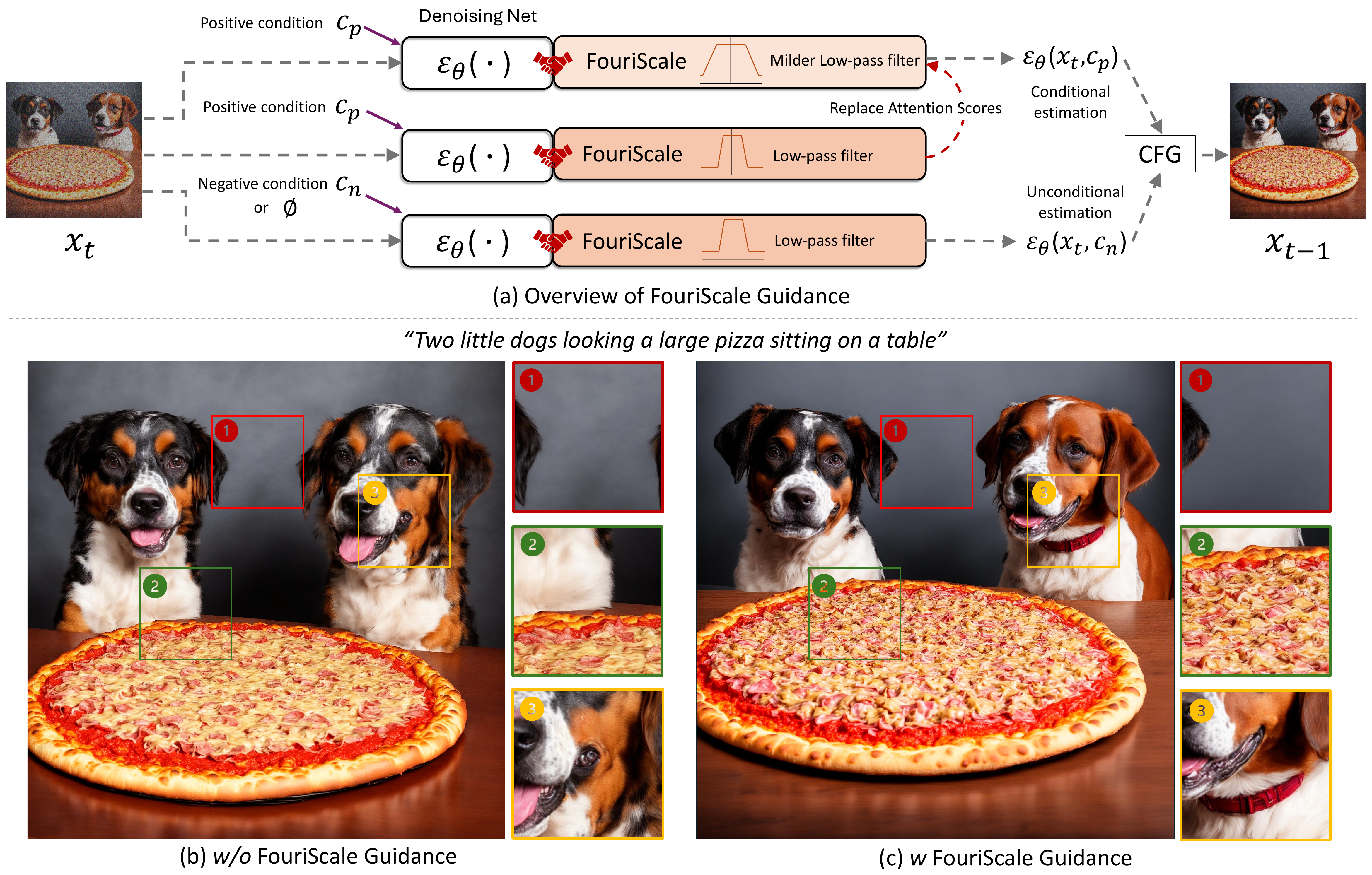}
  \caption{(a) Overview of FouriScale guidance. CFG denotes Classifier-Free Guidance. (b)(c) Visual comparisons between the images created at $2048 \times 2048$ by SD 2.1: (b) without the application of FouriScale guidance, \blackdingB{1}{red} has unexpected artifacts in the background, \blackdingB{2}{teal}\blackdingB{3}{orange} are wrong details, (c) with the application of FouriScale guidance.} \label{fig:fouriscale_guidance}
  \vspace{-3mm}
\end{figure}

\subsection{FouriScale Guidance} \label{sec:fouri_guidance}
FouriScale effectively mitigates structural distortion when generating high-res images. However, it would introduce certain artifacts and unforeseen patterns in the background, as depicted in Fig.~\ref{fig:fouriscale_guidance} (b). Based on our empirical findings, we identify that the main issue stems from the application of low-pass filtering when generating the conditional estimation in classifier-free guidance~\cite{ho2022classifier}. This process often leads to a ringing effect and loss of detail. To improve image quality and reduce artifacts, as shown in Fig.~\ref{fig:fouriscale_guidance} (a), 
we develop a guided version of FouriScale for reference, aiming to align the output, rich in details, with it.
Specifically, beyond the unconditional and conditional estimations derived from the UNet modified by FouriScale, we further generate an extra conditional estimation. This one is subjected to identical dilated convolutions but utilizes milder low-pass filters to accommodate more frequencies. We substitute its attention maps of attention layers with those from the conditional estimation processed through FouriScale, in a similar spirit with image editing~\cite{hertz2022prompt,cao2023masactrl,epstein2023diffusion}. Given that UNet's attention maps hold a wealth of positional and structural information~\cite{zhao2023unleashing,xiao2023text,wang2023diffusion}, this strategy allows for the incorporation of correct structural information derived from FouriScale to guide the generation, simultaneously mitigating the decline in image quality and loss of details typically induced by low-pass filtering. The final noise estimation is determined using both the unconditional and the newly conditional estimations following classifier-free guidance. As we can see in Fig.~\ref{fig:fouriscale_guidance} (c), the aforementioned issues are largely mitigated.

\subsection{Detailed Designs}

\paragraph{Annealing dilation and filtering.}

Since the image structure is primarily outlined in the early reverse steps, the subsequent steps focus on enhancing the details, we implement an annealing approach for both dilation convolution and low-pass filtering. Initially, for the first $S_{init}$ steps, we employ the ideal dilation convolution and low-pass filtering. During the span from $S_{init}$ to $S_{stop}$, we progressively decrease the dilation factor and $r$ (as detailed in Algorithm \ref{algo:final_operation}) down to 1. After $S_{stop}$ steps, the original UNet is utilized to refine image details further.

\paragraph{Settings for SDXL.}

Stable Diffusion XL~\cite{podell2023sdxl} (SDXL) is generally trained on images with a resolution close to $1024 \times 1024$ pixels, accommodating various aspect ratios simultaneously. Our observations reveal that using an ideal low-pass filter leads to suboptimal outcomes for SDXL. Instead, a gentler low-pass filter, which modulates rather than completely eliminates high-frequency elements using a coefficient $\sigma \in [0,1]$ (set to 0.6 in our method) delivers superior visual quality. This phenomenon can be attributed to SDXL's ability to handle changes in scale effectively, negating the need for an ideal low-pass filter to maintain scale consistency, which confirms the rationale of incorporating low-pass filtering to address scale variability. 
Additionally, for SDXL, we calculate the scale factor $r$ (refer to Algorithm \ref{algo:final_operation}) by determining the training resolution whose aspect ratio is closest to the one of target resolution.

\section{Experiments} \label{sec:experiments}

\paragraph{Experimental setup.} 

Wo follow~\cite{he2023scalecrafter} to report results on three text-to-image models, including SD 1.5~\cite{epstein2023diffusion}, SD 2.1~\cite{sd2-1-base}, and SDXL 1.0~\cite{podell2023sdxl} on generating images at four higher resolutions. The resolutions tested are 4$\times$, 6.25$\times$, 8$\times$, and 16$\times$ the pixel count of their respective training resolutions. For both SD 1.5 and SD 2.1 models, the original training resolution is set at 512$\times$512 pixels, while the inference resolutions are 1024$\times$1024, 1280$\times$1280, 2048$\times$1024, and 2048$\times$2048. In the case of the SDXL model, it is trained at resolutions close to 1024$\times$1024 pixels, with the higher inference resolutions being 2048$\times$2048, 2560$\times$2560, 4096$\times$2048, and 4096$\times$4096. We default use FreeU~\cite{si2023freeu} in all experimental settings.

\paragraph{Testing dataset and evaluation metrics.} Following~\cite{he2023scalecrafter}, we assess performance using the Laion-5B dataset~\cite{laion5b}, which comprises 5 billion pairs of images and their corresponding captions. For tests conducted at an inference resolution of 1024$\times$1024, we select a subset of 30,000 images, each paired with randomly chosen text prompts from the dataset. Given the substantial computational demands, our sample size is reduced to 10,000 images for tests at inference resolutions exceeding 1024$\times$1024.
We evaluate the quality and diversity of the generated images by measuring the Frechet Inception Distance (FID)~\cite{heusel2017gans} and Kernel Inception Distance (KID)~\cite{binkowski2018demystifying} between generated images and real images, denoted as FID$_r$ and KID$_r$. To show the methods' capacity to preserve the pre-trained model’s original ability at a new resolution, we also follow~\cite{he2023scalecrafter} to evaluate the metrics between the generated images at the base training resolution and the inference resolution, denoted as FID$_b$ and KID$_b$.

\begin{table}[!t]
    \centering
        \caption{Quantitative comparisons among training-free methods. The best and second best results are highlighted in \textbf{bold} and \underline{underline}. The values of KID$_r$ and KID$_b$ are scaled by $10^2$.}
    \vspace{-3mm}
    \label{tab:quantitative_results}
    \resizebox{0.98\linewidth}{!}{
    \begin{tabular}{lccccccccccccc}
\toprule
\multirow{2}{*}{Resolution} & \multirow{2}{*}{Method} & \multicolumn{4}{c}{SD 1.5} & \multicolumn{4}{c}{SD 2.1} & \multicolumn{4}{c}{SDXL 1.0} \\
\cmidrule(lr){3-6} \cmidrule(lr){7-10} \cmidrule(lr){11-14}
& & $\text{FID}_r\downarrow$ & $\text{KID}_r\downarrow$ & $\text{FID}_b\downarrow$ & $\text{KID}_b\downarrow$ & $\text{FID}_r\downarrow$ & $\text{KID}_r\downarrow$ & $\text{FID}_b\downarrow$ & $\text{KID}_b\downarrow$ & $\text{FID}_r\downarrow$ & $\text{KID}_r\downarrow$ & $\text{FID}_b\downarrow$ & $\text{KID}_b\downarrow$ \\
\midrule
\multirow{4}{*}{4$\times$ 1:1} & Vanilla & 26.96 & 1.00 & 15.72 & 0.42 & 29.90 & 1.11 & 19.21 & 0.54 & 49.81 & 1.84 & 32.90 & 0.92 \\
& Attn-Entro & 26.78 & 0.97 & 15.64 & 0.42 & 29.65 & 1.10 & 19.17 & 0.54 & 49.72 & 1.84 & 32.86 & 0.92 \\
& ScaleCrafter & \underline{23.90} & \underline{0.95} & \underline{11.83} & \underline{0.32} & \underline{25.19} & \textbf{0.98} & \underline{13.88} & \textbf{0.40} & \underline{49.46} & \underline{1.73} & \underline{36.22} & \underline{1.07} \\
& Ours & \textbf{23.62} & \textbf{0.92} & \textbf{10.62} & \textbf{0.29} & \textbf{25.17} & \textbf{0.98} & \textbf{13.57} & \textbf{0.40} & \textbf{33.89} & \textbf{1.21} & \textbf{20.10} & \textbf{0.47} \\
\midrule
\multirow{4}{*}{6.25$\times$ 1:1} & Vanilla & 41.04 & 1.28 & 31.47 & 0.77 & 45.81 & 1.52 & 37.80 & 1.04 & 68.87 & 2.79 & 54.34 & 1.92 \\
& Attn-Entro & 40.69 & 1.31 & 31.25 & 0.76 & 45.77 & 1.51 & 37.75 & 1.04 & 68.50 & 2.76 & 54.07 & 1.91 \\
& ScaleCrafter & \underline{37.71} & \underline{1.34} & \underline{25.54} & \underline{0.67} & \underline{35.13} & \underline{1.14} & \underline{23.68} & \underline{0.57} & \underline{55.03} & \underline{2.02} & \underline{45.58} & \underline{1.49} \\
& Ours & \textbf{30.27} & \textbf{1.00} & \textbf{16.71} & \textbf{0.34} & \textbf{30.82} & \textbf{1.01} & \textbf{18.34} & \textbf{0.42} & \textbf{44.13} & \textbf{1.64} & \textbf{37.09} & \textbf{1.16} \\
\midrule
\multirow{4}{*}{8$\times$ 1:2} & Vanilla & 50.91 & 1.87 & 44.65 & 1.45 & 57.80 & 2.26 & 51.97 & 1.81 & 90.23 & 4.20 & 79.32 & 3.42 \\
& Attn-Entro & 50.72 & 1.86 & 44.49 & 1.44 & 57.42 & 2.26 & 51.67 & 1.80 & \underline{89.87} & \underline{4.15} & \underline{79.00} & \underline{3.40} \\
& ScaleCrafter & \underline{35.11} & \underline{1.22} & \underline{29.51} & \underline{0.81} & \underline{41.72} & \underline{1.42} & \underline{35.08} & \underline{1.01} & 106.57 & 5.15 & 108.67 & 5.23 \\
& Ours & \textbf{35.04} & \textbf{1.19} & \textbf{26.55} & \textbf{0.72} & \textbf{37.19} & \textbf{1.29} & \textbf{27.69} & \textbf{0.74} & \textbf{71.77} & \textbf{2.79} & \textbf{70.70} & \textbf{2.65} \\
\midrule
\multirow{4}{*}{16$\times$ 1:1} & Vanilla & 67.90 & 2.37 & 66.49 & 2.18 & 84.01 & 3.28 & 82.25 & 3.05 & 116.40 & 5.45 & 109.19 & 4.84 \\
& Attn-Entro & 67.45 & 2.35 & 66.16 & 2.17 & 83.68 & 3.30 & 81.98 & 3.04 & 113.25 & 5.44 & 106.34 & 4.81 \\
& ScaleCrafter & \underline{32.00} & \underline{1.01} & \underline{27.08} & \underline{0.71} & \underline{40.91} & \underline{1.32} & \underline{33.23} & \underline{0.90} & \underline{84.58} & \underline{3.53} & \underline{85.91} & \underline{3.39} \\
& Ours & \textbf{30.84} & \textbf{0.95} & \textbf{23.29} & \textbf{0.57} & \textbf{39.49} & \textbf{1.27} & \textbf{28.14} & \textbf{0.73} & \textbf{56.66} & \textbf{2.18} & \textbf{49.59} & \textbf{1.63} \\
\bottomrule
\end{tabular}}
\end{table}

\subsection{Quantitative Results}

We compare our method with the vanilla text-to-image diffusion model (Vanilla), the training-free approach~\cite{jin2023training} (Attn-Entro) that accounts for variations in attention entropy between low and high resolutions, and ScaleCrafter~\cite{he2023scalecrafter}, which modifies convolution kernels through re-dilation and adopts linear transformations for kernel enlargement. We show the experimental results in Tab.~\ref{tab:quantitative_results}. 
Compared to the vanilla diffusion models, our method obtains much better results because of eliminating the issue of repetitive patterns. 
The Attn-Entro does not work at high upscaling levels because it fails to fundamentally consider the structural consistency across resolutions. 
Due to the absence of scale consistency consideration in ScaleCrafter, it performs worse than our method on the majority of metrics. Additionally, we observe that ScaleCrafter often struggles to produce acceptable images for SDXL,  leading to much lower performance than ours.
Conversely, our method is capable of generating images with plausible structures and rich details at various high resolutions, compatible with any pre-trained diffusion models. 

Furthermore, our method achieves better inference speed compared with ScaleCrafter~\cite{he2023scalecrafter}. For example, under the 16$\times$ setting for SDXL, ScaleCrafter takes an average of 577 seconds to generate an image, whereas our method, employing a single NVIDIA A100 GPU, averages 540 seconds per image.

\subsection{Qualitative Results}

\begin{figure}[!t]
  \centering
  \includegraphics[width=0.98\columnwidth]{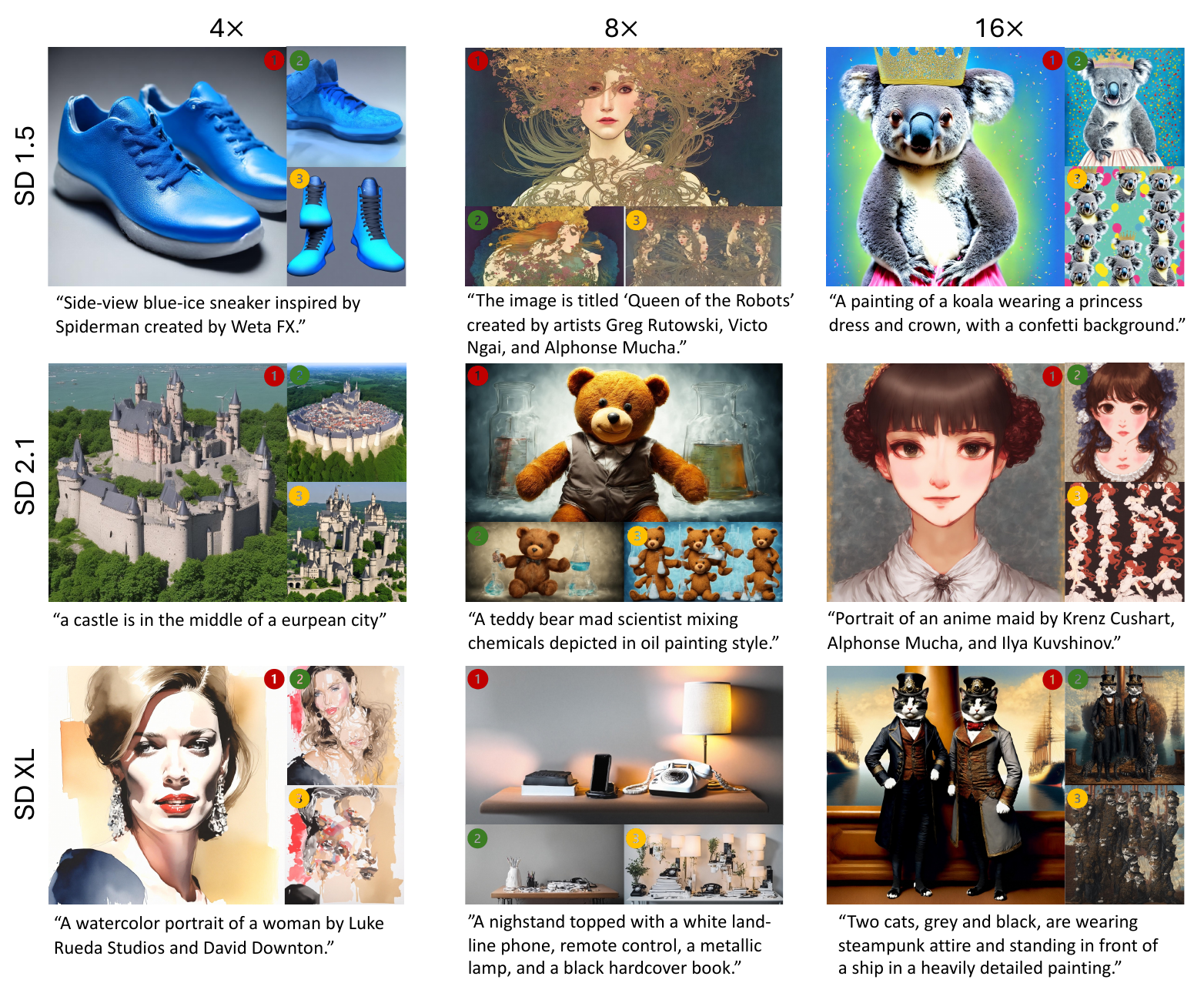}
  \caption{Visual comparisons between \blackdingB{1}{red} ours, \blackdingB{2}{teal} ScaleCrafter~\cite{he2023scalecrafter} and \blackdingB{3}{orange} Attn-Entro~\cite{jin2023training}, under settings of 4$\times$, 8$\times$, and 16$\times$, employing three distinct pre-trained diffusion models: SD 1.5, SD 2.1, and SDXL 1.0.} \label{fig:visualization_main}
\end{figure}

Fig.~\ref{fig:visualization_main} presents a comprehensive visual comparison across various upscaling factors (4$\times$, 8$\times$, and 16$\times$) with different pre-trained diffusion models (SD 1.5, 2.1, and SDXL 1.0). Our method 
demonstrates superior performance in preserving structural integrity and fidelity compared to ScaleCrafter~\cite{he2023scalecrafter} and Attn-Entro~\cite{jin2023training}. Besides, FouriScale maintains its strong performance across all three pre-trained models, demonstrating its broad applicability and robustness.
At 4$\times$ upscaling, FouriScale faithfully reconstructs fine details like the intricate patterns on the facial features of the portrait, and textures of the castle architecture. In contrast, ScaleCrafter and Attn-Entro often exhibit blurring and loss of details.
As we move to more extreme 8$\times$ and 16$\times$ upscaling factors, the advantages of FouriScale become even more pronounced. Our method consistently generates images with coherent global structures and locally consistent textures across diverse subjects, from natural elements to artistic renditions. The compared methods still struggle with repetitive artifacts and distorted shapes.

\subsection{Ablation Study}

To validate the contributions of each component in our proposed method, we conduct ablation studies on the SD 2.1 model generating $2048\times2048$ images.

\begin{figure}[!t]
    \centering
    \begin{minipage}{0.4\textwidth}
        \centering
        \vspace{4mm}
        \begin{tabular}{l|c}
            \toprule
            Method & FID$_r$ \\
            \midrule
            FouriScale & 39.49 \\
            \emph{w/o} guidance & 43.75 \\
            \emph{w/o} guidance \& filtering & 46.74 \\
            \bottomrule
        \end{tabular}
        \captionsetup{singlelinecheck=off}
        \vspace{4mm}
        \captionof{table}{Ablation studies on FouriScale components on SD 2.1 model under $16\times$ 1:1 setting.}
        \label{tab:guidance_ablation}
    \end{minipage}%
    \hfill
    \begin{minipage}{0.58\textwidth}
        \centering
        \includegraphics[width=0.98\linewidth]{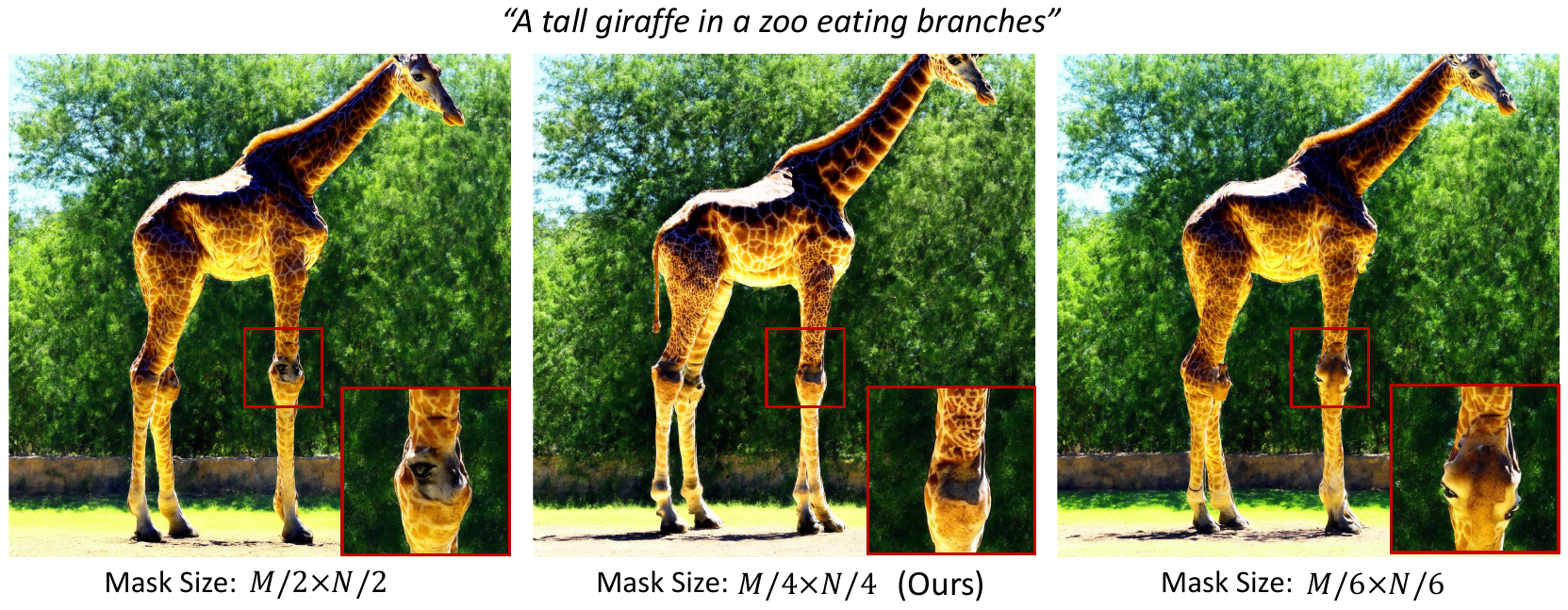}
        \captionsetup{singlelinecheck=off}
        \caption{Comparison of mask sizes for passing low frequencies generating 2048$^2$ images by SD 2.1. $M$, $N$ denote height and width of target resolution.}
        \label{fig:mask_size_comparison}
    \end{minipage}
\end{figure}

First, we analyze the effect of using FouriScale Guidance as described in Sec.~\ref{sec:fouri_guidance}. We compare the default FouriScale which utilizes guidance versus removing the guidance and solely relying on the conditional estimation from the FouriScale-modified UNet. As shown in Tab.~\ref{tab:guidance_ablation}, employing guidance improves the FID$_r$ by 4.26, demonstrating its benefits for enhancing image quality. The guidance allows incorporating structural information from the FouriScale-processed estimation to guide the generation using a separate conditional estimation with milder filtering. This balances between maintaining structural integrity and preventing loss of details.

Furthermore, we analyze the effect of the low-pass filtering operation described in Sec.~\ref{sec:low_pass}. Using the FouriScale without guidance as the baseline, we additionally remove the low-pass filtering from all modules. As shown in Tab.~\ref{tab:guidance_ablation}, this further deteriorates the FID$_r$ to 46.74. The low-pass filtering is crucial for maintaining scale consistency across resolutions and preventing aliasing effects that introduce distortions. Without it, the image quality degrades significantly. 

A visual result of comparing the mask sizes for passing low frequencies is depicted in Fig.~\ref{fig:mask_size_comparison}. The experiment utilizes SD 2.1 (trained with 512$\times$512 images) to generate images of 2048$\times$2048 pixels, setting the default mask size to $M/4 \times N/4$. We can find that the optimal visual result is achieved with our default settings. As the low-pass filter changes, there is an evident deterioration in the visual appearance of details, which underscores the validity of our method.

\section{Conclusion and Limitation}

We present FouriScale, a novel approach that enhances the generation of high-resolution images from pre-trained diffusion models. By addressing key challenges such as repetitive patterns and structural distortions, FouriScale introduces a training-free method based on frequency domain analysis, improving structural and scale consistency across different resolutions by a dilation operation and a low-pass filtering operation. The incorporation of a padding-then-cropping strategy and the application of FouriScale guidance enhance the flexibility and quality of text-to-image generation, accommodating different aspect ratios while maintaining structural integrity. FouriScale's simplicity and adaptability, avoiding any extensive pre-computation, set a new benchmark in the field. FouriScale still faces challenges in generating ultra-high-resolution samples, such as 4096$\times$4096 pixels, which typically exhibit unintended artifacts. Additionally, its focus on operations within convolutions limits its applicability to purely transformer-based diffusion models.

\newpage
\appendix
\section*{Appendix} \label{sec:appendix_A}

\setcounter{section}{0}
\setcounter{table}{0}
\setcounter{figure}{0}
\setcounter{equation}{0}
\renewcommand{\thetable}{\arabic{table}}
\renewcommand{\thefigure}{\arabic{figure}}

\section{Proof}

\subsection{Proof of Theorem \ref{theo:1d_signal}} \label{sec:proof_theorem}

Let's consider $f(x)$ as a one-dimensional signal. Its down-sampled counterpart is represented by $f'(x) = \operatorname{Down}_s(f)$. To understand the connection between $f'(x)$ and $f(x)$, we base our analysis on their generated continuous signal $g(x)$, which is produced using a particular sampling function. It's important to note that the sampling function $sa_{\Delta T}(x)$ is characterized by a series of infinitely spaced impulse units, with each pair separated by intervals of $\Delta T$:
\begin{equation} \label{eq:1d_sampling}
    sa(x, \Delta T) = \sum_{n=-\infty}^{\infty} \delta (x - n\Delta T).
\end{equation}

Based on Eq.~\eqref{eq:1d_sampling}, $f(x)$ and $f'(x)$ can be formulated as
\begin{equation}
\begin{aligned}
f(x) &= g(x) sa(x, \Delta T), \\
f'(x) &= g(x) sa(x, s\Delta T).
\end{aligned}
\end{equation}

Based on the Fourier transform and the convolution theorem, the spatial sampling described above can be represented in the Fourier domain as follows:
\begin{flalign} \label{eq:freq_sampling}
    F(u) &= G(u) \circledast SA(u, \Delta T) \nonumber \\
    &= \int_{-\infty}^{\infty} G(\tau)SA(u - \tau, \Delta T)d\tau \nonumber \\
&= \frac{1}{\Delta T} \sum_{n} \int_{-\infty}^{\infty} G(\tau) \delta \left( u - \tau - \frac{n}{\Delta T} \right) d\tau \\
& = \frac{1}{\Delta T} \sum_{n} G \left( u - \frac{n}{\Delta T} \right), \nonumber
\end{flalign}
where $G(u)$ and $SA(u, \Delta T)$ are the Fourier transform of $g(x)$ and $sa(x, \Delta T)$. From the above Equation, it can be observed that the spatial sampling introduces the periodicity to the spectrum and the period is \(\frac{1}{\Delta T}\).

Note that the sampling rates of \(f(x)\) and \(f'(x)\) are \(\Omega_x\) and \(\Omega'_x\), the relationship between them can be written as
\begin{equation}
\Omega_x = \frac{1}{\Delta T}, \quad \Omega'_x = \frac{1}{s\Delta T} = \frac{1}{s}\Omega_x.
\end{equation}

With the down-sampling process in consideration, we presume that \(f(x)\) complies with the Nyquist sampling theorem, suggesting that \(u_{max} < \frac{\Omega_x}{2}\).

Following down-sampling, as per the Nyquist sampling theorem, the entire sub-frequency range is confined to $(0, \frac{\Omega_x}{s})$. 
The resulting frequency band is a composite of $s$ initial bands, expressed as:
\begin{equation} \label{eq:s_superpose}
   F'(u) = \mathbb{S}(F(u), F(\tilde{u}_1), \ldots, F(\tilde{u}_{s-1})), 
\end{equation}
where \(\tilde{u}_i\) represents the frequencies higher than the sampling rate, while \(u\) denotes the frequencies that are lower than the sampling rate. The symbol \(\mathbb{S}\) stands for the superposition operator. To simplify the discussion, \(\tilde{u}\) will be used to denote \(\tilde{u}_i\) in subsequent sections.

(1) In the sub-band, where \( u \in (0, \frac{\Omega_x}{2s}) \), \(\tilde{u}\) should satisfy
\begin{equation} \label{eq:aliasing_theorem1}
   \quad \tilde{u} \in \left(\frac{\Omega_x}{2s}, u_{max}\right). 
\end{equation}

According to the aliasing theorem, the high frequency \(\tilde{u}\) is folded back to the low frequency:
\begin{equation} \label{eq:aliasing_theorem2}
\hat{u} = \left| \tilde{u} - (k + 1)\frac{\Omega'_x}{2} \right|, \quad k\frac{\Omega'_x}{2} \leq \tilde{u} \leq (k + 2)\frac{\Omega'_x}{2}
\end{equation}
where \(k = 1, 3, 5, \ldots\) and \(\hat{u}\) is folded results by \(\tilde{u}\).

According to Eq.~\ref{eq:aliasing_theorem1} and Eq.~\ref{eq:aliasing_theorem2}, we have
\begin{equation} \label{eq:u_hat_range}
\hat{u} = \frac{a\Omega_x}{s} - \tilde{u} \quad \text{and} \quad \hat{u} \in \left(\frac{\Omega_x}{s} - u_{max}, \frac{\Omega_x}{2s}\right),
\end{equation}
where \(a = (k+1) / 2 = 1, 2, \ldots\). According to Eq.~\eqref{eq:s_superpose} and Eq.~\eqref{eq:u_hat_range}, we can attain
\begin{equation} \label{eq:fu_case}
F'(u) = 
\begin{cases}
F(u) & \text{if } u \in (0, \frac{\Omega_x}{s} - u_{max}), \\
\mathbb{S}(F(u), F(\frac{a\Omega_x}{s} - u)) & \text{if } u \in (\frac{\Omega_x}{s} - u_{max}, \frac{\Omega_x}{2s}).
\end{cases}
\end{equation}

According to Eq.~\eqref{eq:freq_sampling}, \(F(u)\) is symmetric with respect to \(u = \frac{\Omega_x}{2}\):
\begin{equation} \label{eq:symmetry}
F(\frac{\Omega_x}{2} - u) = F(u + \frac{\Omega_x}{2}).
\end{equation}
Therefore, we can rewrite $F(\frac{a\Omega_x}{s} - u)$ as:
\begin{equation} \label{eq:symmetry_transfer}
\begin{aligned} 
&F(\frac{\Omega_x}{2} - (\frac{\Omega_x}{2}+u-\frac{a\Omega_x}{s})) \\
= &F(\frac{\Omega_x}{2} + (\frac{\Omega_x}{2}+u-\frac{a\Omega_x}{s})) \\
= &F(u + \Omega_x -\frac{a\Omega_x}{s}) \\
= &F(u + \frac{a\Omega_x}{s})
\end{aligned}
\end{equation}
since \(a = 1, 2, \ldots, s-1\). Additionally, for \(s = 2\), the condition \(u \in (0, \frac{\Omega_x}{s} - u_{max})\) results in \(F(u + \frac{\Omega_x}{s}) = 0\). When \(s > 2\), the range \(u \in (0, \frac{\Omega_x}{s} - u_{max})\) typically becomes non-existent. Thus, in light of Eq.~\eqref{eq:symmetry_transfer} and the preceding analysis, Eq.~\eqref{eq:fu_case} can be reformulated as
\begin{equation} \label{eq:theorem_prove1}
F'(u) = \mathbb{S}(F(u), F(u + \frac{a\Omega_x}{s})) \mid u \in (0, \frac{\Omega_x}{2s}).
\end{equation}

(2) In the sub-band, where \(u \in (\frac{\Omega_x}{2s}, \frac{\Omega_x}{s})\), different from (1), \(\tilde{u}\) should satisfy
\begin{equation}
\tilde{u} \in (\frac{\Omega_x}{s} - u_{max}, \frac{\Omega_x}{2s}).
\end{equation}

Similarly, we can obtain:
\begin{equation} \label{eq:theorem_prove2}
F'(u) = \mathbb{S}(F(\tilde{u}), F(u + \frac{a\Omega_x}{s})) \mid u \in (\frac{\Omega_x}{2s}, \frac{\Omega_x}{s}).
\end{equation}

Combining Eq.~\eqref{eq:theorem_prove1} and Eq.~\eqref{eq:theorem_prove2}, we obtain
\begin{equation}
F'(u) = \mathbb{S}(F(u), F(u + \frac{a\Omega_x}{s})) \mid u \in (0, \frac{\Omega_x}{s}),
\end{equation}
where \(a = 1, 2, \ldots, s-1\).

\subsection{Proof of Lemma \ref{lemma:image_superpose}} \label{sec:proof_lemma}

Based on Eq.~\eqref{eq:freq_sampling}, it can be determined that the amplitude of \( F' \) is \(\frac{1}{s}\) times that of \( F \). Hence, \( F'(u) \) can be expressed as:
\begin{equation}
F'(u) = \frac{1}{s}F(u) + \sum_a \frac{1}{s}F\left(u + \frac{a\Omega_x}{s}\right) \mid u \in \left(0, \frac{\Omega_x}{s}\right).
\end{equation}

Based on the dual principle, we can prove \( F'(u, v) \) in the whole sub-band
\begin{equation}
    F'(u,v) = \frac{1}{s^2} \left(\sum_{a,b=0}^{s-1} F\left(u + \frac{a\Omega_s}{s}, v + \frac{b\Omega_y}{s} \right)\right),
\end{equation}
where \( u \in \left(0, \frac{\Omega_x}{s}\right) \), \( v \in \left(0, \frac{\Omega_y}{s}\right) \).

\section{Implementation Details}
\subsection{Low-pass Filter Definition}

\begin{figure}[!t]
  \centering
  \includegraphics[width=0.98\columnwidth]{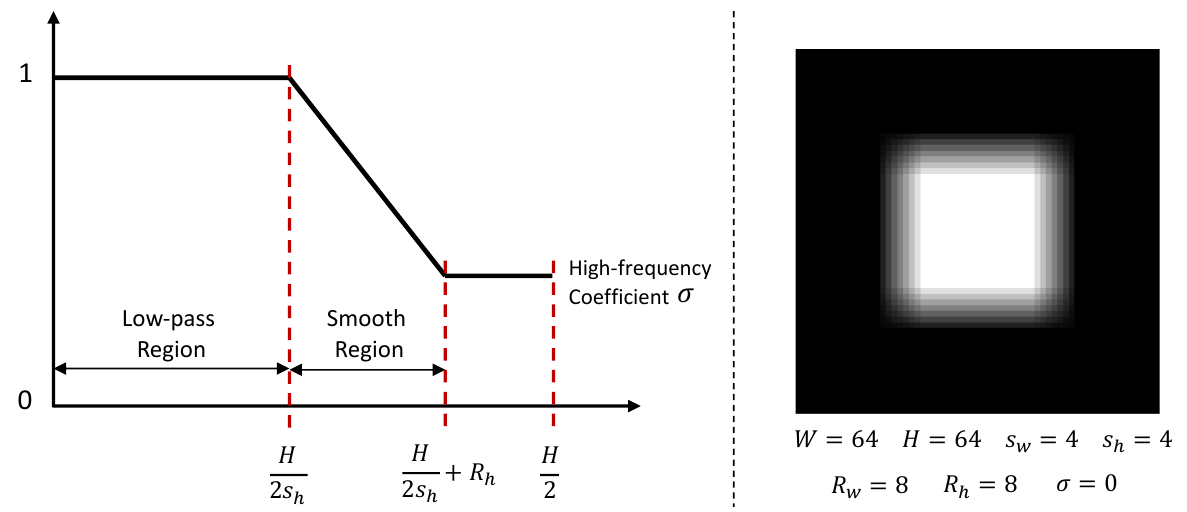}
  \caption{Visualization of the design of a low-pass filter. (a) 1D filter for the positive axis. (2) 2D low-pass filter, which is constructed by mirroring the 1D filters and performing an outer product between two 1D filters, in accordance with the settings of the 1D filter.} \label{fig:filter_vis}
\end{figure}

In Fig.~\ref{fig:filter_vis}, we show the design of a low-pass filter used in FouriScale. Inspired by~\cite{sukhbaatar2019adaptive,riad2021learning}, we define the low-pass filter as the outer product between two 1D filters (depicted in the left of Fig.~\ref{fig:filter_vis}), one along the height dimension and one along the width dimension. We define the function of the 1D filter for the height dimension as follows, filters for the width dimension can be obtained in the same way:
\begin{equation}
    \text{mask}^h_{(s_{h},R_h,\sigma)} = \min \left( \max \left( \frac{1 - \sigma}{R_h} \left( \frac{H}{s_{h}} + 1 - i \right) + 1, \sigma \right), 1 \right), i \in [0,\frac{H}{2}],
\end{equation}
where $s_h$ denotes the down-sampling factor between the target and original resolutions along the height dimension. $R_h$ controls the smoothness of the filter and $\sigma$ is the modulation coefficient for high frequencies. Exploiting the characteristic of conjugate symmetry of the spectrum, we only consider the positive axis, the whole 1D filter can be obtained by mirroring the 1D filter. We build the 2D low-pass filter as the outer product between the two 1D filters:
\begin{equation}
    \text{mask}(s_{h}, s_{w}, R_h, R_w, \sigma) = 
\text{mask}^h_{(s_{h},R_h,\sigma)} \otimes \text{mask}^w_{(s_{w},R_w,\sigma)},
\end{equation}
where $\otimes$ denotes the outer product operation. Likewise, the whole 2D filter can be obtained by mirroring along the height and width axes. A toy example of a 2D low-pass filter is shown in the right of Fig.~\ref{fig:filter_vis}.

\subsection{Hyper-parameter Settings}

\begin{figure}[t!]
    \centering
    \includegraphics[width=0.9\columnwidth]{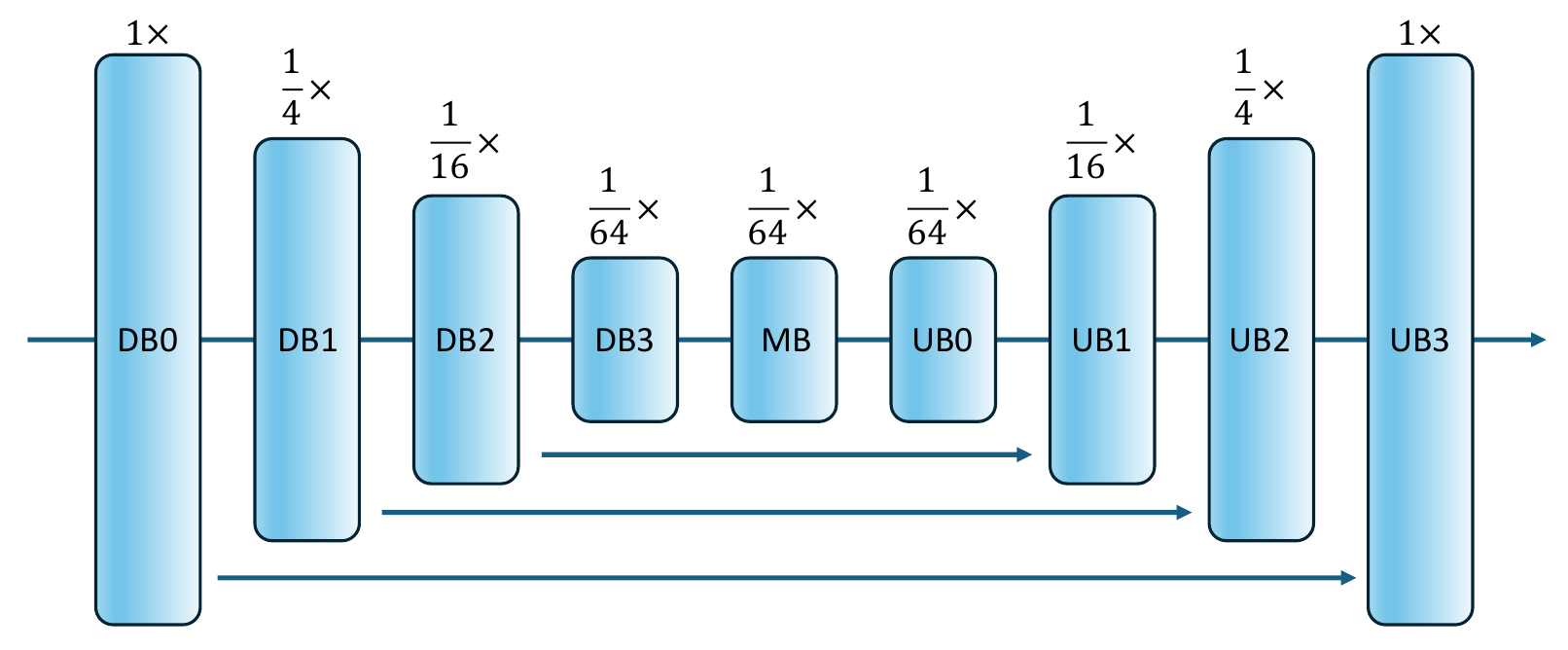}
    \caption{Reference block names of stable diffusion in the following experiment details.}
    \label{fig:architecture}
    \vspace{-5mm}
\end{figure}

In this section, we detail our choice of hyperparameters. The evaluative parameters are detailed in Tab.~\ref{tab:layer_setting}. Additionally, Fig.~\ref{fig:architecture} provides a visual guide of the precise positioning of various blocks within the U-Net architecture employed in our model. 

The dilation factor used in each FouriScale layer is determined by the maximum value of the height and width scale relative to the original resolution. As stated in our main manuscript, we employ an annealing strategy. For the first $S_{init}$ steps, we employ the ideal dilation convolution and low-pass filtering. During the span from $S_{init}$ to $S_{stop}$, we progressively decrease the dilation factor and $r$ (as detailed in Algorithm 1 of our main manuscript) down to 1. After $S_{stop}$ steps, the original UNet is utilized to refine image details further. The settings for $S_{init}$ and $S_{stop}$ are shown in Tab.~\ref{tab:layer_setting}.

\begin{table}[h]
    \centering
    \caption{Experiment settings for SD 1.5, SD 2.1, and SDXL 1.0.}
    \label{tab:layer_setting}
    \begin{tabular}{ccc}
        \toprule
        Params & SD 1.5 \& SD 2.1 & SDXL 1.0  \\
        \midrule
        FouriScale blocks & [DB2,DB3,MB,UB0,UB1,UB2] & [DB2,MB,UB0,UB1] \\
        inference timesteps & 50 & 50 \\
        \multirow{2}{*}{[$S_{init}$, $S_{stop}$]} & [10,30] (4$\times$1:1 and 6.25$\times$1:1) & \multirow{2}{*}{[20,35]}  \\
        & [20,35] (8$\times$1:2 and 16$\times$1:1) \\
        \bottomrule
    \end{tabular}
\end{table}

\section{More Experiments}

\subsection{Comparison with Diffusion Super-Resolution Method}

\begin{figure}[t!]
    \centering
    \includegraphics[width=0.92\columnwidth]{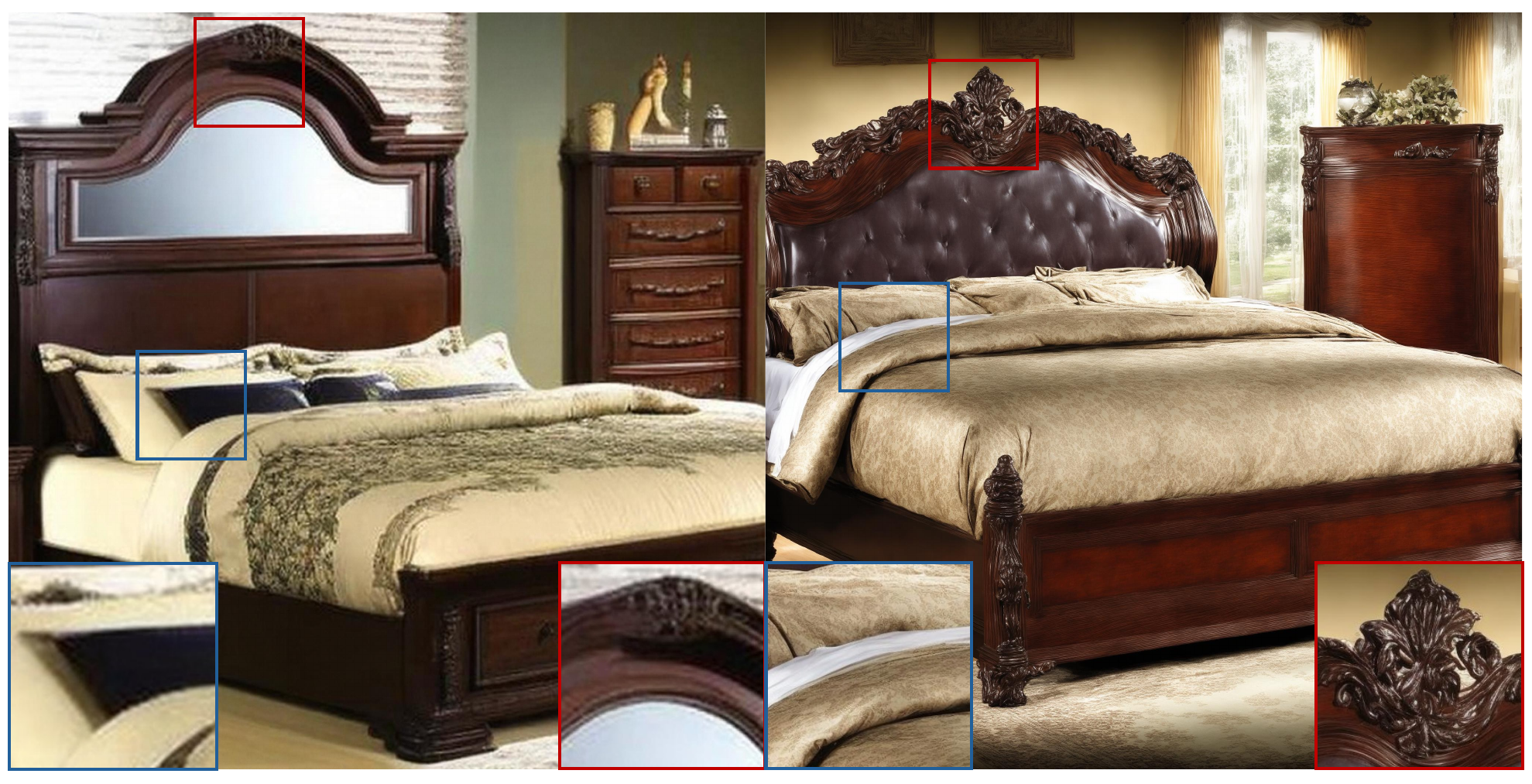}
    \caption{Visual comparison with SD+SR. \textbf{Left:} 2048$\times$2048 image upscaled by SD+SR from 512$\times$512 SD 2.1 generated image. \textbf{Right:} 2048$\times$2048 image generated by our FouriScale with SD 2.1.}
    \label{fig:sd_sr}
\end{figure}

In this section, we compare the performance of our proposed method with a cascaded pipeline, which uses SD 2.1 to generate images at the default resolution of 512$\times$512, and upscale them to 2048$\times$2048 by a
pre-trained diffusion super-resolution model, specifically the Stable Diffusion Upscaler-4$\times$~\cite{x4-upscaler}. We apply this super-resolution model to a set of 10,000 images generated by SD 2.1. We then evaluate the FID$_r$ and KID$_r$ scores of these upscaled images and compare them with images generated at 2048$\times$2048 resolution using SD 2.1 equipped with our FouriScale. The results of this comparison are presented in Tab.~\ref{tab:sd_sr}. 
Our method obtains somewhat worse results than the cascaded method. However, our method is capable of generating high-resolution images in only one stage, without the need for a multi-stage process. Besides, our method does not need model re-training, while the SR model demands extensive data and computational resources for training. More importantly, as shown in Fig.~\ref{fig:sd_sr}, we find that our method can generate much better details than the cascaded pipeline. Due to a lack of prior knowledge in generation, super-resolution method can only utilize existing knowledge within the single image for upscaling the image, resulting in over-smooth appearance. However, our method can effectively upscale images and fill in details using generative priors with a pre-trained diffusion model, providing valuable insights for future explorations into the synthesis of high-quality and ultra-high-resolution images.

\begin{figure}[tb!]
    \centering
    \begin{minipage}{0.38\textwidth}
        \centering
        \captionof{table}{Comparison with SD + Super-Resolution.}
        \label{tab:sd_sr}
        \begin{tabular}{lcc}
            \toprule
            Method & FID$_r$ & KID$_r$ \\
            \midrule
            SD+Super Resolution & 25.94 & 0.91 \\
            Ours & 39.49 & 1.27 \\
            \bottomrule
        \end{tabular}
    \end{minipage}%
    \hfill
    \begin{minipage}{0.52\textwidth}
        \centering
        \captionof{table}{Comparison with ElasticDiffusion on the SDXL 2048$\times$2048 setting.}
        \vspace{1mm}
        \label{tab:elastic}
        \begin{tabular}{lcccc}
        \toprule
        Method & FID$_r$ & KID$_r$ & FID$_b$ & KID$_b$ \\
        \midrule
        ElasticDiffusion \cite{haji2023elasticdiffusion} & 52.02 & 3.03 & 40.46 & 2.22 \\
        Ours & 33.89 & 1.21 & 20.10 & 0.47 \\
        \bottomrule
        \end{tabular}
    \end{minipage}
\end{figure}

\begin{figure}[!t]
  \centering
  \includegraphics[width=0.98\columnwidth]{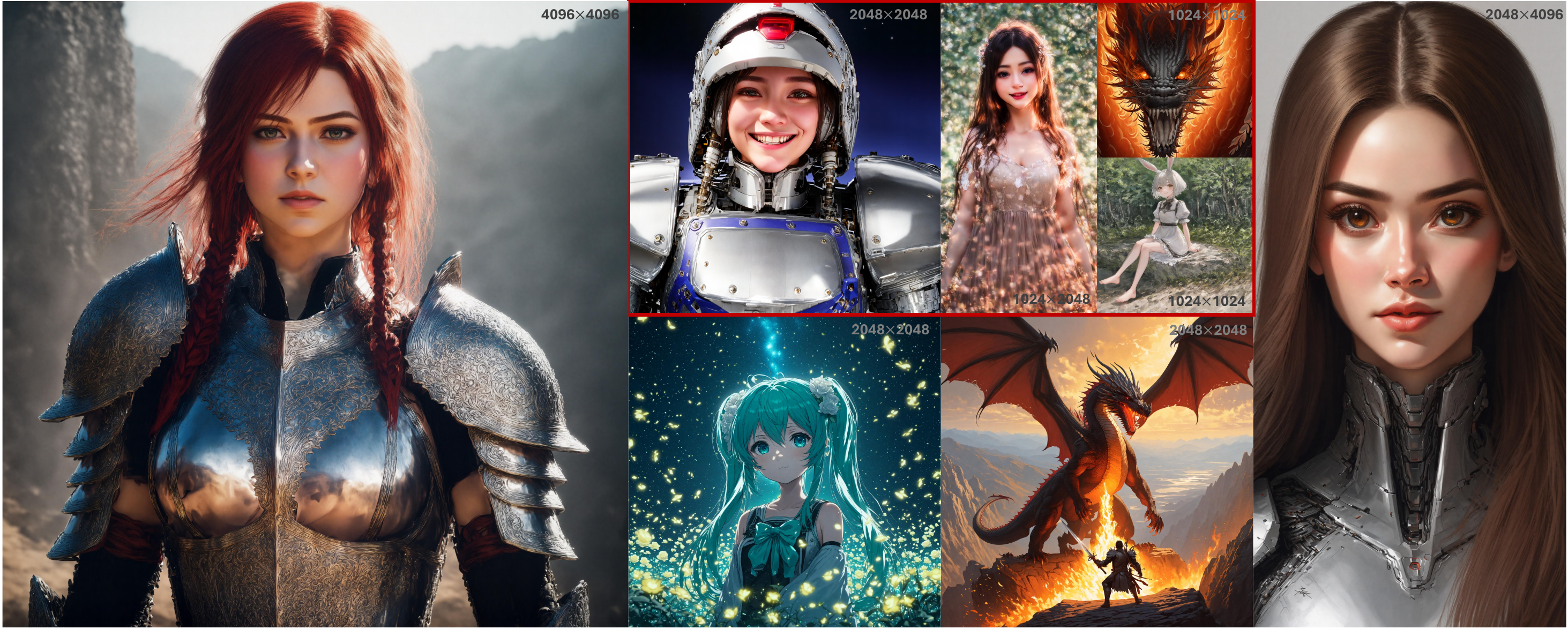}
  \caption{Visualization of the high-resolution images generated by SD 2.1 integrated with customized LoRAs (images in red rectangle) and images generated by a personalized diffusion model, AnimeArtXL~\cite{animeartxl}, which is based on SDXL.} \label{fig:LoRA}
\end{figure}

\begin{figure}[!h]
  \centering
\includegraphics[width=0.8\columnwidth]{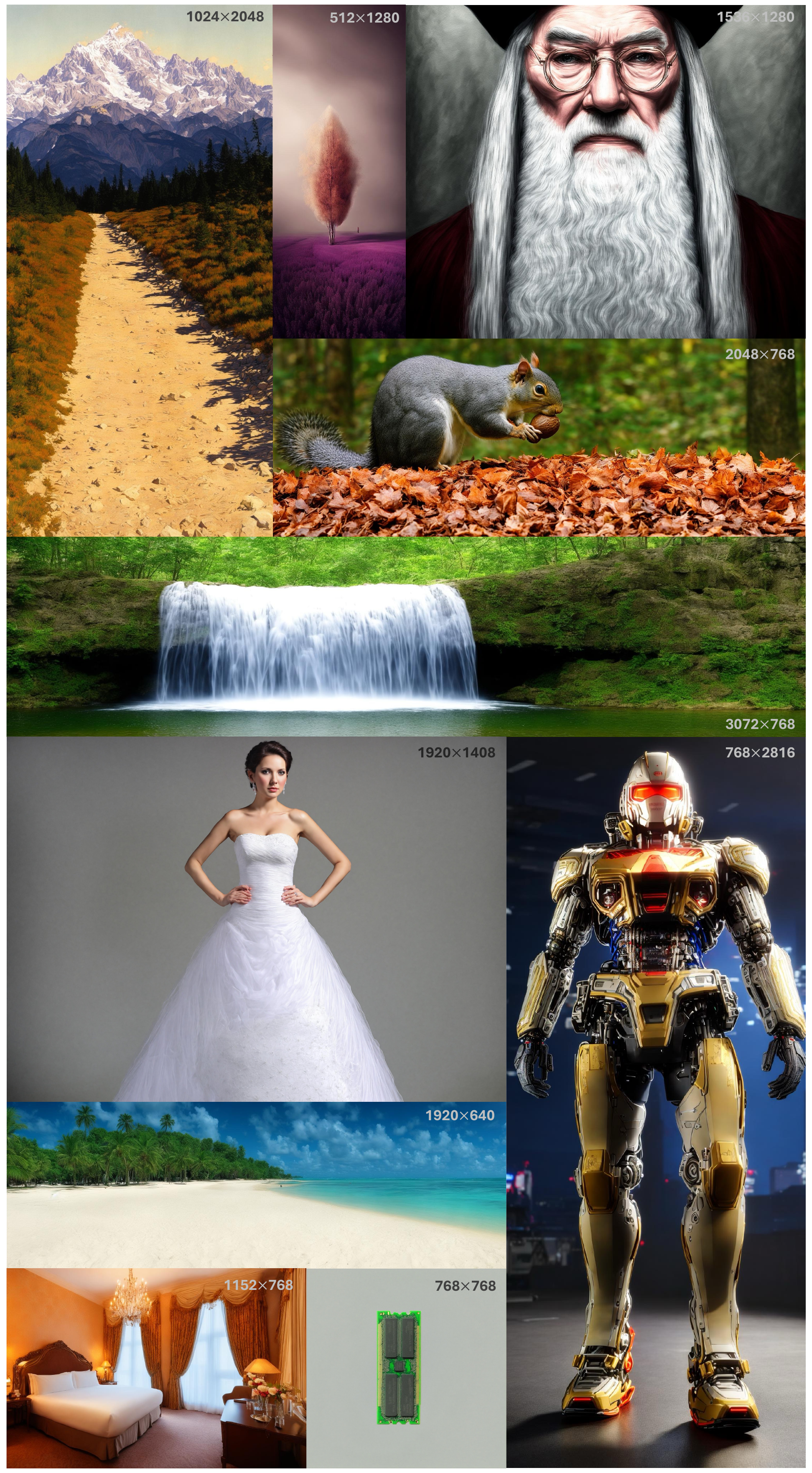}
  \caption{More generated images using FouriScale and SD 2.1 with arbitrary resolutions.} \label{fig:more_visualization}
\end{figure}

\subsection{Comparison with ElasticDiffusion}

We observe that the recent approach, ElasticDiffusion~\cite{haji2023elasticdiffusion}, has established a technique to equip pre-trained diffusion models with the capability to generate images of arbitrary sizes, both smaller and larger than the resolution used during training. Here, we provide the comparison with ElasticDiffusion \cite{haji2023elasticdiffusion} on the SDXL 2048$\times$2048 setting. The results are shown in Tab~\ref{tab:elastic}. First, it's important to note that the inference times for ElasticDiffusion are approximately 4 to 5 times longer than ours. Besides, as we can see, our method demonstrates superior performance across all evaluation metrics, achieving lower FID and KID scores compared to ElasticDiffusion, indicating better image quality and diversity. 

\section{More Visualizations}

\subsection{LoRAs}
In Fig.~\ref{fig:LoRA}, we present the high-resolution images produced by SD 2.1, which has been integrated with customized LoRAs~\cite{hu2021lora} from Civitai~\cite{civitai}. We can see that our method can be effectively applied to diffusion models equipped with LoRAs.

\subsection{Other Resolutions}
In Fig.~\ref{fig:more_visualization}, we present more images generated at different resolutions by SD 2.1, aside from the 4$\times$, 6.25$\times$, 8$\times$, and 16$\times$ settings.
Our approach is capable of generating high-quality images of arbitrary aspect ratios and sizes.

\bibliographystyle{splncs04}
\bibliography{main}
\end{document}